\RequirePackage{fix-cm}
\documentclass[twocolumn]{svjour3}          
\smartqed  
\usepackage{graphicx}
\usepackage[square,sort,comma,numbers]{natbib}
\usepackage{multirow}
\usepackage{amsmath}
\usepackage{amssymb}
\usepackage{bbm}
\usepackage{bm}
\usepackage{booktabs}
\usepackage{array} 
\usepackage{blindtext}
\usepackage{comment}

\usepackage{pifont}

\usepackage[dvipsnames]{xcolor}
\usepackage{color, colortbl}
\definecolor{citecolor}{HTML}{0071bc}
\definecolor{tabhighlight}{HTML}{e5e5e5}
\usepackage[colorlinks,citecolor=citecolor]{hyperref}

\newcommand{\keypoint}[1]{\noindent\textbf{#1}\quad}
\makeatletter
\DeclareRobustCommand\onedot{\futurelet\@let@token\@onedot}
\def\@onedot{\ifx\@let@token.\else.\null\fi\xspace}
\def\eg{\emph{e.g}\onedot, } 
\def\ie{\emph{i.e}\onedot, } 
\def\cf{\emph{c.f}\onedot} 
\def\etc{\emph{etc}\onedot} \def\vs{\emph{vs}~}
\def\wrt{w.r.t\onedot} 

\makeatother

\newcommand{\revise}[1]{{\color{black}{#1}}}

\makeatletter
\renewcommand\paragraph{
  \@startsection{paragraph} 
  {4} 
  {\z@} 
  {.5em \@plus1ex \@minus.2ex} 
  {-.5em} 
  {\normalfont\normalsize\bfseries} 
}
\makeatother
\begin{document}
\sloppy

\title{Generalized Out-of-Distribution Detection: A Survey 
}


\author{Jingkang Yang \and
        Kaiyang Zhou \and
        Yixuan Li \and
        Ziwei Liu
}


\institute{Jingkang Yang \at
              S-Lab, Nanyang Technological University, Singapore \\
              \email{jingkang001@ntu.edu.sg}
           \and
           Kaiyang Zhou \at
              S-Lab, Nanyang Technological University, Singapore \\
              \email{kaiyang.zhou@ntu.edu.sg}
           \and
           Yixuan Li \at
              Department of Computer Sciences, University of Wisconsin-Madison, Madison, WI, United States\\
              \email{sharonli@cs.wisc.edu}
           \and
           Ziwei Liu \at
           S-Lab, Nanyang Technological University, Singapore \\
           \email{ziwei.liu@ntu.edu.sg}
}

\date{Received: date / Accepted: date}

\maketitle

\begin{abstract}
Out-of-distribution (OOD) detection is critical to ensuring the reliability and safety of machine learning systems. 
For instance, in autonomous driving, we would like the driving system to issue an alert and hand over the control to humans when it detects unusual scenes or objects that it has never seen during training time and cannot make a safe decision. 
The term, OOD detection, first emerged in 2017 and since then has received increasing attention from the research community, leading to a plethora of methods developed, ranging from classification-based to density-based to distance-based ones. 
Meanwhile, several other problems, including anomaly detection (AD), novelty detection (ND), open set recognition (OSR), and outlier detection (OD), are closely related to OOD detection in terms of motivation and methodology.
Despite common goals, these topics develop in isolation, and their subtle differences in definition and problem setting often confuse readers and practitioners.
In this survey, we first present a unified framework called \emph{generalized OOD detection}, which encompasses the five aforementioned problems, \ie AD, ND, OSR, OOD detection, and OD. 
Under our framework, these five problems can be seen as special cases or sub-tasks, and are easier to distinguish. 
Despite comprehensive surveys of related fields, the summarization of OOD detection methods remains incomplete and requires further advancement. 
This paper specifically addresses the gap in recent technical developments in the field of OOD detection. It also provides a comprehensive discussion of representative methods from other sub-tasks and how they relate to and inspire the development of OOD detection methods. The survey concludes by identifying open challenges and potential research directions.
\end{abstract}

\section{Introduction}\label{sec:introduction}
A trustworthy visual recognition system should not only produce accurate predictions on known context, but also detect unknown examples and reject them (or hand them over to human users for safe handling)~\cite{concrete16arxiv,mlsafety21arxiv,hendrycks2021unsolved,hendrycks2022x}. 
For instance, a well-trained food classifier should be able to detect non-food images such as selfies uploaded by users, and reject such input instead of blindly classifying them into existing food categories. 
In safety-critical applications such as autonomous driving, the driving system must issue a warning and hand over the control to drivers when it detects unusual scenes or objects it has never seen during training.

\begin{figure}[h]
\vspace{-8mm}
    \centering
    \includegraphics[width=\linewidth]{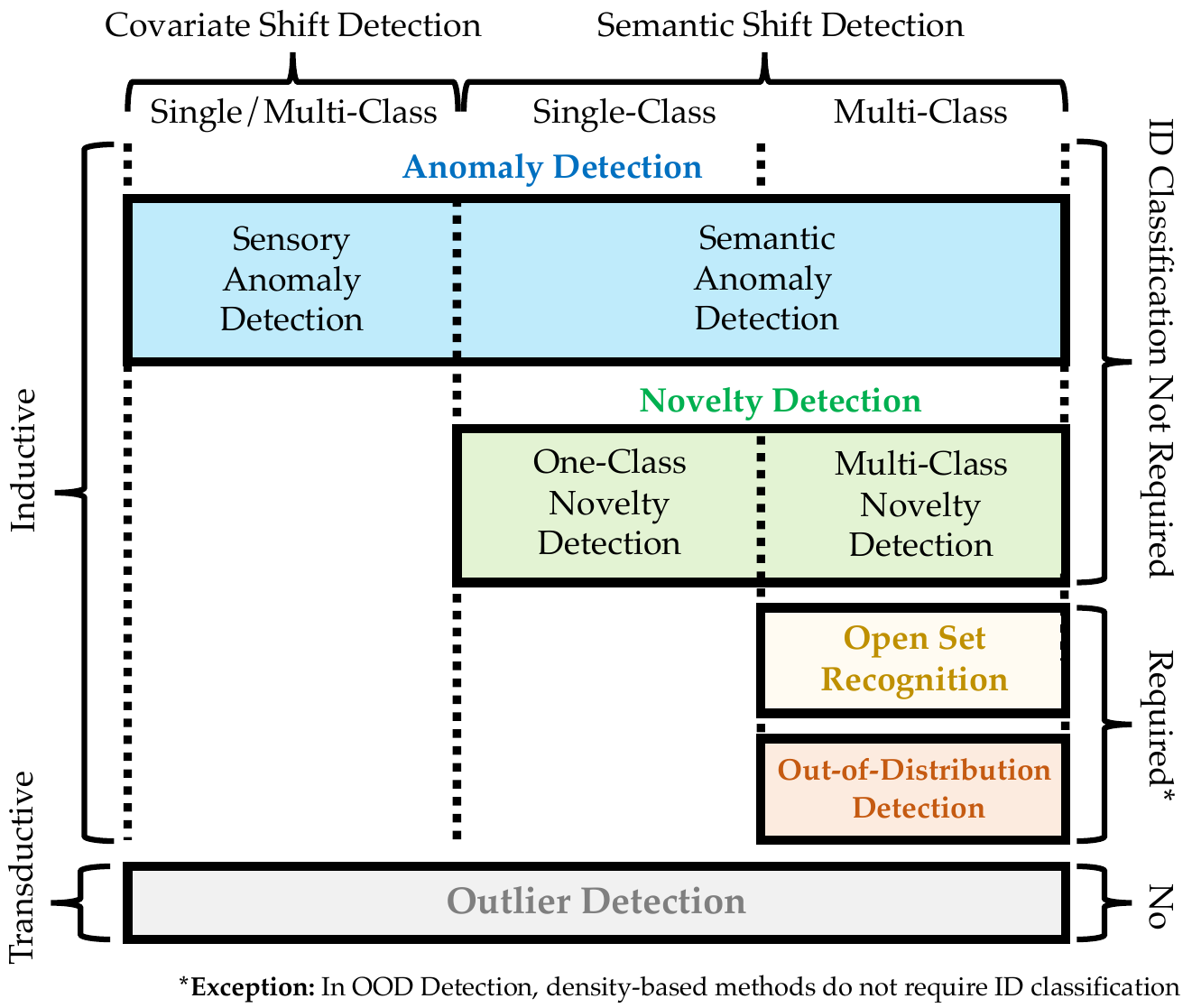}
    \caption{Taxonomy of generalized OOD detection framework, illustrated by classification tasks.
    Four bases are used for the task taxonomy: 
    \textbf{1)} Distribution shift to detect: the task focuses on detecting covariate shift or semantic shift; 
    \textbf{2)} ID data type: the ID data contains one single class or multiple classes; 
    \textbf{3)} Whether the task requires ID classification;
    \textbf{4)} Transductive learning task requires all observations;  inductive tasks follow the train-test scheme.
    Note that ND is often interchangeable with AD, but ND is more concerned with semantic anomalies. 
    OOD detection is generally interchangeable with OSR for classification tasks.}
    \vspace{-3mm}
    \label{fig:taxonomy}
\end{figure}

\begin{figure*}[t]
    \centering
    \includegraphics[width=\textwidth]{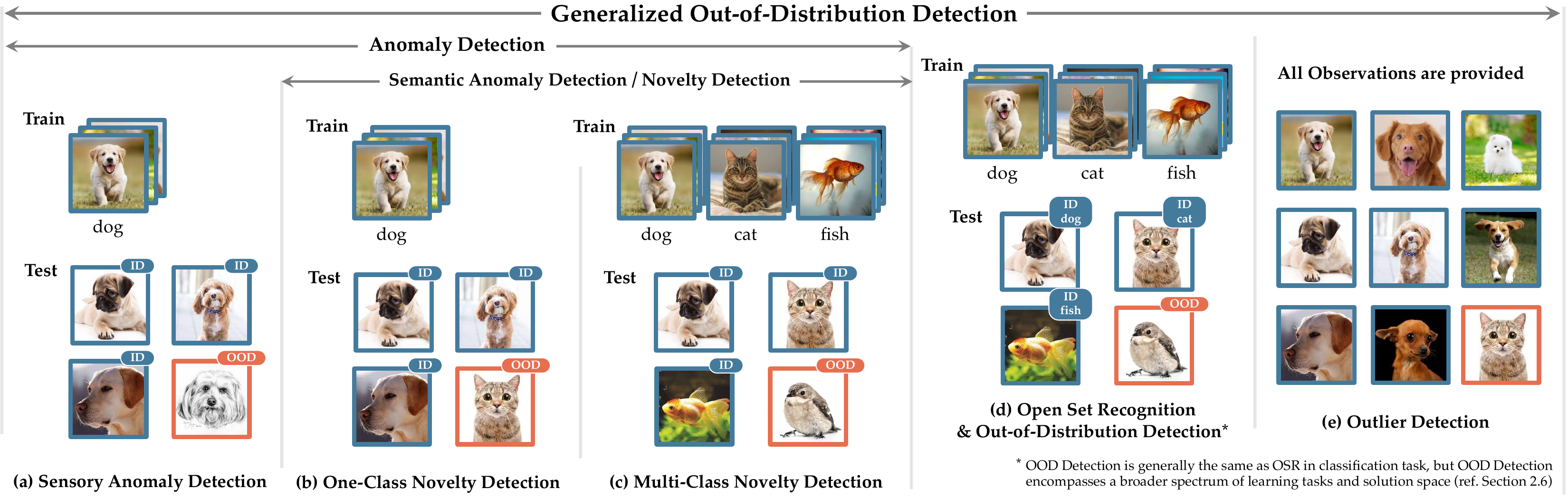}
    \caption{Illustration of sub-tasks under generalized OOD detection framework with vision tasks.
    Tags on test images refer to model's expected predictions.
    \textbf{(a)}~In \emph{sensory anomaly detection}, test images with covariate shift will be considered as OOD. No semantic shift occurs in this setting.
    \textbf{(b)}~In \emph{one-class novelty detection}, normal/ID images belong to one class. Test images with semantic shift will be considered as OOD.
    \textbf{(c)}~In \emph{multi-class novelty detection}, ID images belong to multiple classes. Test images with semantic shift will be considered as OOD.
    Note that \textbf{(b)} and \textbf{(c)} compose novelty detection, which is identical to the topic of semantic anomaly detection.
    \textbf{(d)}~\emph{Open set recognition} is identical to multi-class novelty detection in the task of detection, with the only difference that open set recognition further requires ID classification.
    \emph{Out-of-distribution detection} solves the same problem as open-set recognition. It canonically aims to detect test samples with semantic shift without losing the ID classification accuracy.
    However, OOD Detection encompasses a broader spectrum of learning tasks and solution space.
    \textbf{(e)} \emph{Outlier detection} does not follow a train-test scheme. All observations are provided. It fits in the generalized OOD detection framework by defining the majority distribution as ID. Outliers can have any distribution shift from the majority.
    }
    \label{fig:benchmark}
\end{figure*}

Most existing machine learning models are trained based on the closed-world assumption~\cite{imagenet12nips,surpass15iccv}, where the test data is assumed to be drawn \emph{i.i.d.} from the same distribution as the training data, known as in-distribution (ID). However, when models are deployed in an \emph{open-world} scenario~\cite{openworld06esi}, test samples can be out-of-distribution (OOD) and therefore should be handled with caution. 
The distributional shifts can be caused by semantic shift (\eg OOD samples are drawn from different classes)~\cite{oodbaseline17iclr}, or covariate shift (\eg OOD samples from a different domain)~\cite{domainshift10ml,deepdg17iccv,dasurvey18neurocomp}.

The detection of semantic distribution shift (\eg due to the occurrence of new classes) is the focal point of OOD detection tasks, where the label space $\mathcal{Y}$ can be different between ID and OOD data and hence the model should not make any prediction. In addition to OOD detection, several problems adopt the ``open-world'' assumption and have a similar goal of identifying OOD examples. These include outlier detection~(OD)~\cite{outlierhighd01sigmod,outliersurvey04aireview,outlier05handbook,outlierprogress19ieee}, anomaly detection~(AD)~\cite{anomalysurvey21ieee,anomalyreview20adelaide,anomalysurvey20dsong,anomalysurvey19sydney}, novelty detection~(ND)~\cite{ndsurveyox14sp,ndreview10mipro,ndsurvey03sp01,ndsurvey03sp02}, and open set recognition~(OSR)~\cite{boult19aaai,osrsurvey20pami,osrsurvey21arxiv}. 
While all these problems are related to each other by sharing similar motivations, subtle differences exist among the \emph{sub-topics} in terms of the specific definition. 
However, the lack of a comprehensive understanding of the relationship between the different sub-topics leads to confusion for both researchers and practitioners. Even worse, these sub-topics, which are supposed to be compared and learned from each other, are developing in isolation.

In this survey, we for the first time clarify the similarities and differences between these problems, and present a unified framework termed \emph{generalized OOD detection}.
Under this framework, the five problems (\ie AD, ND, OSR, OOD detection, and OD) can be viewed as special cases or sub-topics. 
While other sub-topics have been extensively surveyed, the summarization of OOD detection methods is still inadequate and requires further exploration. This paper fills this gap by focusing specifically on recent technical developments in OOD detection, analyzing fair experimental comparisons among classical methods on common benchmarks. Our survey concludes by highlighting open challenges and outlining potential avenues for future research.

We further conduct a literature review for each sub-topic, with a special focus on the OOD detection task. To sum up, we make three contributions to the research community:
\begin{enumerate}
\item \textbf{A Unified Framework}:
For the first time, we systematically review five closely related topics of AD, ND, OSR, OOD detection, and OD, and present a unified framework of \emph{generalized OOD detection}. Under this framework, the similarities and differences of the five sub-topics can be systematically compared and analyzed. We hope our unification helps the community better understand these problems and correctly position their research in the literature.

\item \textbf{A Comprehensive Survey for OOD Detection}:
Noticing the existence of comprehensive surveys on AD, ND, OSR, and OD methodologies in recent years~\cite{anomalysurvey21ieee,anomalyreview20adelaide,anomalysurvey20dsong,anomalysurvey19sydney,osrsurvey20pami}, this survey provides a comprehensive overview of OOD detection methods and thus complements existing surveys.
By connecting with methodologies of other sub-topics that are also briefly reviewed, as well as sharing the insights from a fair comparison on a standard benchmark, we hope to provide readers with a more holistic understanding of the developments for each problem and their interconnections, especially for OOD detection.

\item \textbf{Future Research Directions}:
We draw readers' attention to some problems or limitations that remain in the current generalized OOD detection field. We conclude this survey with discussions on open challenges and opportunities for future research.
\end{enumerate}
\section{Generalized OOD Detection}
\label{sec:general-ood}
\keypoint{Framework Overview} In this section, we introduce a unified framework termed \emph{generalized OOD detection}, which encapsulates five related sub-topics: anomaly detection~(AD), novelty detection~(ND), open set recognition~(OSR), out-of-distribution~(OOD) detection, and outlier detection~(OD).
These sub-topics can be similar in the sense that they all define a certain \emph{in-distribution}, with the common goal of detecting \emph{out-of-distribution} samples under the open-world assumption.
However, subtle differences exist among the sub-topics in terms of the specific definition and properties of ID and OOD data---which are often overlooked by the research community. To this end, we provide a clear introduction and description of each sub-topic in respective subsections (from Section~\ref{sec:tax_anomaly} to \ref{sec:tax_od}). Each subsection details the motivation, background, formal definition, as well as relative position within the unified framework. 
Applications and benchmarks are also introduced, with concrete examples that facilitate understanding.
Fig.~\ref{fig:benchmark} illustrates the settings for each sub-topic.
In the end, we conclude this section by introducing the neighborhood topics to clarify the scope of the generalized OOD detection framework. (Section~\ref{sec:related_topics}).

\keypoint{Preliminary: Distribution Shift} 
\revise{
In our framework, we recognize the complexity and interconnectedness of distribution shifts, which are central to understanding various OOD scenarios. Distribution shifts can be broadly categorized into \textit{covariate shift} and \textit{semantic (label) shift}, but it's important to clarify their interdependence.
Firstly, let's define the input space as $\mathcal{X}$ (sensory observations) and the label space as $\mathcal{Y}$ (semantic categories). The data distribution is represented by the joint distribution $P(X, Y)$ over the space $\mathcal{X} \times \mathcal{Y}$.
Distribution shift can occur in either the marginal distribution $P(X)$, or both $P(Y)$ and $P(X)$. Note that shift in $P(Y)$ naturally triggers shift in $P(X)$.

\keypoint{Covariate Shift:} This occurs when there is a change in the marginal distribution $P(X)$, affecting the input space, while the label space $\mathcal{Y}$ remains constant. 
Examples of covariate distribution shift on $P(X)$ include adversarial examples~\cite{adversarial15iclr,adversarial18iclr}, domain shift~\cite{quinonero2009dataset}, and style changes~\cite{gatys2016image}.

\keypoint{Semantic Shift:} This involves changes in both $P(Y)$ and indirectly $P(X)$. A shift in the label space $P(Y)$ implies the introduction of new categories or the alteration of existing ones. This change naturally affects the input space $P(X)$ since the nature of the data being observed or collected is now different. 

\keypoint{Remark:} Given the interdependence between $P(X)$ and $P(Y)$, it's crucial to distinguish the intentions behind different types of distribution shifts. We define \textit{Covariate Shift} as scenarios where changes are intended in the input space ($P(X)$) without any deliberate alteration to the label space ($P(Y)$). On the other hand, \textit{Semantic Shift} specifically aims to modify the semantic content, directly impacting the label space ($P(Y)$) and, consequently, the input space ($P(X)$).

Importantly, we note that covariate shift is more commonly used to evaluate model \emph{generalization} and robustness performance, where the label space $\mathcal{Y}$ remains the same during test time. On the other hand, the detection of semantic distribution shift (\eg due to the occurrence of new classes) is the focal point of many \emph{detection} tasks considered in this framework, where the label space $\mathcal{Y}$ can be different between ID and OOD data and hence the model should not make any prediction.
}

With the concept of distribution shift in mind, readers can get a general idea of the differences and connections among sub-topics/tasks in Fig.~\ref{fig:taxonomy}. 
Notice that different sub-tasks can be easily identified with the following four dichotomies: 
1) covariate/semantic shift dichotomy; 
2) single/multiple class dichotomy; 
3) ID classification needed/non-needed dichotomy; 
4) inductive/transductive dichotomy.
Next, we proceed with elaborating on each sub-topic.

\subsection{Anomaly Detection}
\label{sec:tax_anomaly}
\keypoint{Background}
The notion of ``anomaly'' stands in contrast with the ``normal'' defined in advance.
The concept of ``normal'' should be clear and reflect the real task.
For example, to create a ``not-hotdog detector", the concept of the normal should be clearly defined as the hotdog class, \ie a food category, so that objects that violate this definition are identified as anomalies, which include steaks, rice, and non-food objects like cats and dogs.
Ideally, ``hotdog'' would be regarded as a homogeneous concept, regardless of the sub-classes of French or American hotdog.

Current anomaly detection settings often restrict the environment of interest to some specific scenarios.
For example, the ``not-hotdog detector'' only focuses on realistic images,
assuming the nonexistence of images from other domains such as sketches.
Another realistic example is industrial defect detection, which is based on only one set of assembly lines for a specific product.
In other words, the ``open-world'' assumption is usually not completely ``open".
Nevertheless, ``not-hotdog'' or ``defects'' can form a large unknown space that breaks the ``closed-world'' assumption.

In summary, the key to anomaly detection is to define normal clearly (usually without sub-classes) and detect all possible anomalous samples under some specific scenarios.

\smallskip
\keypoint{Definition}
Anomaly detection (AD)~\cite{chandola2009anomaly} aims to detect any anomalous samples that deviate from the predefined normality during testing. The deviation can happen due to either covariate shift or semantic shift, which leads to two sub-tasks: sensory AD and semantic AD, respectively~\cite{anomalysurvey21ieee}.

Sensory AD detects test samples with covariate shift, under the assumption that normalities come from the same covariate distribution. No semantic shift takes place in sensory AD settings.
On the other hand, semantic AD detects test samples with label shift, assuming that normalities come from the same semantic distribution (category), \ie normalities should belong to only one class.

Formally, in sensory AD, normalities are from in-distribution $P(X)$ while anomalies encountered at test time are from out-of-distribution $P'(X)$, where $P(X) \neq P'(X)$ --- only covariate shift occurs.
The goal in sensory AD is to detect samples from $P'(X)$.
No semantic shift occurs in this setting, \ie $P(Y)=P'(Y)$. Conversely, for semantic AD, only semantic shift occurs (\ie $P(Y) \neq P'(Y)$) and the goal is to detect samples that belong to novel classes.

\smallskip
\keypoint{Remark: Sensory/Semantic Dichotomy}
Our sensory/semantic dichotomy for the AD sub-task definition comes from the low-level sensory anomalies and high-level semantic anomalies that are introduced in ~\cite{semanticanomaly20aaai} and highlighted in the recent AD survey~\cite{anomalysurvey21ieee}, to reflect the rise of deep learning. Note that although most sensory and semantic AD methods are shown to be mutually inclusive due to the common shift on $P(X)$, some approaches are specialized in one of the sub-tasks (ref.~Section~\ref{sec:anomaly}).
Recent research communities are also trending on subdividing types of anomalies to develop targeted methods, so that practitioners can select the optimal solution for their own practical problem~\cite{semanticanomaly20aaai,zhang2021understanding}.

\smallskip
\keypoint{Position in Framework}
Under the generalized OOD detection framework, the definition of ``normality'' seamlessly connects to the notion of ``in-distribution", and ``anomaly'' corresponds to ``out-of-distribution".
Importantly, AD treats ID samples as a whole, which means that regardless of the number of classes (or statistical modalities) in ID data, AD does not require differentiation in the ID samples. This feature is an important distinction between AD and other sub-topics such as OSR and OOD detection.


\smallskip
\keypoint{Application and Benchmark}
Sensory AD only focuses on objects with the same or similar semantics, and identifies the observational differences on their surface. Samples with sensory differences are recognized as sensory anomalies. 
Example applications include adversarial defense~\cite{akhtar2018threat}, forgery recognition of biometrics and artworks~\cite{spoofsmartphone16,facespoof15,spoofscheme08,forgeryart09}, image forensics~\cite{deepfakedataset19,deeperforensics20,forensicsurvey20}, industrial inspection~\cite{mvtec19cvpr,rlad20eccv,healthmonitor18health}, \etc.
The most popular academic AD benchmark is MVTec-AD~\cite{mvtec19cvpr} for industrial inspection.

In contrast to sensory AD, semantic AD only focuses on the semantic shift. 
An example of real-world applications is crime surveillance~\cite{idrees2018enhancing,surveillance02ijcnn}.
Active image crawlers for a specific category also need semantic AD methods to ensure the purity of the collected images~\cite{optimol10ijcv}.
An example of the academic benchmarks is to recursively use one class from MNIST as ID during training, and ask the model to distinguish it from the rest of the 9 classes during testing.

\smallskip
\keypoint{Evaluation}
In the AD benchmarks, test samples are annotated to be either normal or abnormal. 
The deployed anomaly detector will produce a confidence score for a test sample, indicating how confident the model considers the sample as normality.
Samples below the predefined confidence threshold are considered abnormal.
By viewing the anomalies as positive and
true normalities as negative\footnote{Align with MSP~\cite{msp17iclr}. Check \href{https://github.com/Jingkang50/OpenOOD/issues/206}{this issue} in OpenOOD}, different thresholds will produce a series of true positive rates (TPR) and false-positive rates (FPR)---from which we can calculate the area under the receiver operating characteristic curve (AUROC)~\cite{auroc06pr}. Similarly, the precision and recall values can be used to compute metrics of F-scores and the area under the precision-recall curve (AUPR)~\cite{evaluation20jmlt}.
Note that there can be two variants of AUPR values: one treating ``normal'' as the positive class, and the other treating ``abnormal'' as the positive class. For AUROC and AUPR, a higher value indicates better detection performance.

\smallskip
\keypoint{Remark: Alternative Taxonomy on Anomalies}
Some previous literature considers anomalies types to be three-fold: point anomalies, conditional or contextural anomalies, and group or collective anomalies~\cite{anomalyreview20adelaide,anomalysurvey19sydney,anomalysurvey21ieee}.
In this survey, we mainly focus on point anomalies detection for its popularity in practical applications and its adequacy to elucidate the similarities and differences between sub-tasks.
Details of the other two kinds of anomalies, \ie contextural anomalies that often occur in time-series tasks, and collective anomalies that are common in the data mining field, are not covered in this survey. We recommend readers to the recent AD survey papers~\cite{anomalysurvey21ieee} for an in-depth discussion on them.

\smallskip
\keypoint{Remark: Taxonomy based on Supervision}
We use sensory/semantic dichotomy to subdivide AD at the task level. From the perspective of methodologies, some literature categorizes AD techniques into unsupervised and (semi-) supervised settings. Note that these two taxonomies are orthogonal as they focus on tasks and methods respectively.

\subsection{Novelty Detection}
\label{sec:tax_novelty}
\keypoint{Background}
The word ``novel'' generally refers to the unknown, new, and something interesting.
While novelty detection (ND) is often interchangeable with AD in the community, strictly speaking, their subtle difference is worth noticing.
In terms of motivation, novelty detection usually does not perceive ``novel'' test samples as erroneous, fraudulent, or malicious as AD does, but cherishes them as learning resources for potential future use with a positive learning attitude~\cite{anomalyreview20adelaide,anomalysurvey21ieee}. In fact, novelty detection is also known as ``novel class detection''~\cite{ndsurvey03sp01,ndsurvey03sp02}, indicating that it is primarily focusing on detecting semantic shift.

\smallskip
\keypoint{Definition}
Novelty detection aims to detect any test samples that do not fall into any training category.
The detected novel samples are usually prepared for future constructive procedures, such as more specialized analysis, or incremental learning of the model itself.
Based on the number of training classes, ND contains two different settings:
1) one-class novelty detection~(\emph{one-class ND}): only one class exists in the training set;
2) multi-class novelty detection~(\emph{multi-class ND}): multiple classes exist in the training set. It is worth noting that despite having many ID classes, the goal of multi-class ND is only to distinguish novel samples from ID. Both one-class and multi-class ND are formulated as binary classification problems.

\smallskip
\keypoint{Position in Framework}
Under the generalized OOD detection framework, ND deals with the setting where OOD samples have semantic shift, without the need for classification in the ID set even if possible.
Therefore, ND shares the same problem definition with semantic AD.

\smallskip
\keypoint{Application and Benchmark}
Real-world ND application includes video surveillance~\cite{idrees2018enhancing,surveillance02ijcnn}, planetary exploration~\cite{mars19aaai} and incremental learning~\cite{incremental15aims,curiositydriven17icml}.
For one-class ND, an example academic benchmark can be identical to that of semantic AD, which considers one class from MNIST as ID and the rest as the novel.
The corresponding MNIST benchmark for multi-class ND may use the first 6 classes during training, and test on the remaining 4 classes as OOD.

\smallskip
\keypoint{Evaluation}
The evaluation of ND is identical to AD, which is based on AUROC, AUPR, or F-scores (see details in Section~\ref{sec:tax_anomaly}). 

\smallskip
\keypoint{Remark: One-Class/Multi-Class Dichotomy} Although the ND models do not require the ID classification even with multi-class annotations, the method on multi-class ND can be different from one-class ND, as multi-class ND can make use of the multi-class classifier while one-class ND cannot.
Also note that semantic AD can be further split into one-class semantic AD and multi-class semantic AD that matches ND, as semantic AD is equivalent to ND.

\smallskip
\keypoint{Remark: Nuance between AD and ND}
Apart from the special interest in semantics, some literature~\cite{ocgan19cvpr,xia2015learning} also point out that ND is supposed to be fully unsupervised (no novel data in training), while AD might have some abnormal training samples. \revise{It's important to note that neither AD nor ND necessitates the classification of ID data. This is a key distinction between OSR and OOD detection, which we will discuss in subsequent sections.}

\subsection{Open Set Recognition}
\label{sec:tax_osr}
\keypoint{Background}
Machine learning models trained in the closed-world setting can incorrectly classify test samples from unknown classes as one of the known categories with high confidence~\cite{towardosr13pami}.
Some literature refers to this notorious overconfident behavior of the model as ``arrogance", or ``agnostophobia"~\cite{agnostophobia18nips}.
Open set recognition (OSR) is proposed to address this problem, with their own terminology of ``known known classes'' to represent the categories that exist at training, and ``unknown unknown classes'' for test categories that do not fall into any training category. Some other terms, such as open category detection~\cite{pac18icml} and open set learning~\cite{bound21icml}, are simply different expressions for OSR.

\smallskip
\keypoint{Definition}
Open set recognition requires the multi-class classifier to simultaneously: 1) accurately classify test samples from ``known known classes", and 2) detect test samples from ``unknown unknown classes".

\smallskip
\keypoint{Position in Framework}
OSR well aligns with our generalized OOD detection framework, where ``known known classes'' and ``unknown unknown classes'' correspond to ID and OOD respectively.
Formally, OSR deals with the case where OOD samples during testing have semantic shift, \ie $P(Y)\neq P'(Y)$.
The goal of OSR is largely shared with that of multi-class ND---the only difference is that OSR additionally requires accurate classification of ID samples from $P(Y)$.

\smallskip
\keypoint{Application and Benchmark}
OSR supports the robust deployment of real-world image classifiers in general, which can reject unknown samples in the open world~\cite{sorio2010open,openworld19www}.
An example academic benchmark on MNIST can be identical to multi-class ND, which considers the first 6 classes as ID and the remaining 4 classes as OOD. In addition, OSR further requires a good classifier on the 6 ID classes.

\smallskip
\keypoint{Evaluation}
Similar to AD and ND, the metrics for OSR include F-scores, AUROC, and AUPR. Beyond them, the classification performance is also evaluated by standard ID accuracy. While the above metrics evaluate the novelty detection and ID classification capabilities independently, some works raise some evaluation criteria for joint evaluation, such as CCR@FPR$x$~\cite{agnostophobia18nips}, which calculates the class-wise recall when a certain FPR equal to $x$ (\eg $10^{-1}$) is achieved.

\subsection{Out-of-Distribution Detection}
\label{sec:tax_ood}

\keypoint{Background}
With the observation that deep learning models are often inappropriate but in fact overconfident in classifying samples from different semantic distributions in the image classification task and text categorization~\cite{msp17iclr}, the field of out-of-distribution detection emerges, requiring the model to reject inputs that are semantically different from the training distribution and therefore should not be predicted by the model.

\smallskip
\keypoint{Definition}
Out-of-distribution detection, or OOD detection, aims to detect test samples drawn from a distribution that is different from the training distribution, with the definition of distribution to be well-defined according to the application in the target.
For most machine learning tasks, the distribution should refer to ``label distribution'', which means that OOD samples should not have overlapping labels \wrt training data.
Formally, in the OOD detection, the test samples come from a distribution whose semantics are shifted from ID, \ie $P(Y)\neq P'(Y)$.
Note that the training set usually contains multiple classes, and OOD detection should NOT harm the ID classification capability.

\smallskip
\keypoint{Position in Framework}
Out-of-distribution detection can be canonical to OSR in common machine learning tasks like multi-class classification---keeping the classification performance on test samples from ID class space $\mathcal{Y}$, and reject OOD test samples with semantics outside the support of $\mathcal{Y}$. Also, the multi-class setting and the requirement of ID classification distinguish the task from AD and ND.

\smallskip
\keypoint{Application and Benchmark}
The application of OOD detection usually falls into safety-critical situations, such as autonomous driving~\cite{huang2020survey,geiger2012we}.
An example academic benchmark is to use CIFAR-10 as ID during training and to distinguish CIFAR images from other datasets such as SVHN, \etc. 
Researchers should pay attention that OOD datasets should NOT have label overlapping with ID datasets when building the benchmark.

\smallskip
\keypoint{Evaluation}
Apart from F-scores, AUROC, and AUPR, another commonly-used metric is FPR@TPR$x$, which measures the FPR  when the TPR is $x$ (\eg 0.95). Some works also use an alternative metric, TNR@TPR$x$, which is equivalent to 1-FPR@TPR$x$. OOD detection also concerns the performance of ID classification.

\smallskip
\keypoint{Remark: OSR \vs OOD Detection}
The difference between OSR and OOD detection tasks is three-fold.

\noindent\textbf{1) Different benchmark setup:} OSR benchmarks usually split one multi-class classification dataset into ID and OOD parts according to classes, while OOD detection takes one dataset as ID and finds several other datasets as OOD with the guarantee of non-overlapping categories between ID/OOD datasets. 
However, despite the different benchmark traditions of the two sub-tasks, they are in fact tackling the same problem of semantic shift detection.

\noindent\textbf{2) No additional data in OSR:} Due to the requirement of theoretical open-risk bound guarantee, OSR discourages the usage of additional data during training by design~\cite{boult19aaai}. This restriction precludes methods that are more focused on effective performance improvements (\eg outlier exposures~\cite{oe18nips,Zhang_2023_WACV}) but may violate OSR constraints.

\noindent\textbf{3) Broadness of OOD detection:} Compare to OSR, OOD detection encompasses a broader spectrum of learning tasks (\eg multi-label classification~\cite{hendrycks2019scaling}), wider solution space (to be discussed in Section~\ref{sec:ood}).

\smallskip
\keypoint{Remark: Mainstream OOD Detection Focuses on Semantics}
While most works in the current community interpret the keyword ``out-of-distribution'' as ``out-of-label/semantic-distribution'', some OOD detection works also consider detecting covariate shifts~\cite{godin20cvpr}, which claim that covariate shift usually leads to a significant drop in model performance and therefore needs to be identified and rejected.
However, although detecting covariate shift is reasonable on some specific tasks (usually due to high-risk or privacy reasons) that are to be discussed in the following paragraph, research on this topic remains a controversial task \emph{w.r.t} OOD generalization tasks (\cf Section~\ref{sec:related_topics} and Section~\ref{sec:future_direction}). Detecting semantic shift has been the mainstream of OOD detection tasks.

\smallskip
\keypoint{Remark: To Generalize, or To Detect?}
We provide another definition from the perspective of generalization:
Out-of-distribution detection, or OOD detection, aims to detect test samples to which the model cannot or does not want to generalize~\cite{pleiss2019neural}.
In most of the machine learning tasks, such as image classification, the models are expected to generalize their prediction capability to samples with covariate shift, and they are only unable to generalize when semantic shift occurs.
However, for applications where models are by-design nontransferable to other domain, such as many deep reinforcement learning tasks like game AI~\cite{vinyals2017starcraft,sedlmeier2019uncertainty}, the key term ``distribution'' should refer to ``data/input distribution'', so that the model should refuse to decide the environment that is not the same as the training environment, \ie $P(X)\neq P'(X)$. 
Similar applications are those high-risk tasks such as medical image classification~\cite{zimmerer2022mood} or in privacy-sensitive scenario~\cite{tariq2020review}, where the models are expected to be very conservative and only make predictions for samples exactly from the training distribution, rejecting any samples that deviate from it.
Recent studies~\cite{averly2023unified} also highlight a model-specific view: a robust model should generalize to examples with covariate shift; a weak model should reject them.
Ultimately, an OOD detection task is considered valid when it successfully balances the aspects of ``detection" and ``generalization", taking into account factors such as meaningfulness and the inherent challenges presented by the task. Nonetheless, detecting semantic shift remains the primary focus of OOD detection tasks and is central to this survey.

\subsection{Outlier Detection}
\label{sec:tax_od}
\keypoint{Background}
According to~\emph{Wikipedia} \cite{outlierwiki}, an outlier is a data point that differs significantly from other observations.
Recall that the problem settings in AD, ND, OSR, and OOD detect unseen test samples that are different from the training data distribution.
In contrast, outlier detection directly processes all observations and aims to select outliers from the contaminated dataset~\cite{outlier05handbook,outliersurvey04aireview,outlierhighd01sigmod}. Since outlier detection does not follow the train-test procedure but has access to all observations, approaches to this problem are usually transductive rather than inductive~\cite{transductive16iaai}.

\smallskip
\keypoint{Definition}
Outlier detection aims to detect samples that are markedly different from the others in the given observation set, due to either covariate or semantic shift.

\smallskip
\keypoint{Position in Framework}
Different from all previous sub-tasks, whose in-distribution is defined during training, the ``in-distribution'' for outlier detection refers to the majority of the observations. Outliers may exist due to semantic shift on $P(Y)$, or covariate shift on $P(X)$.

\smallskip
\keypoint{Application and Benchmark}
While mostly applied in data mining tasks~\cite{ben2005outlier,basu2007automatic,download19asonam}, outlier detection is also used in real-world computer vision applications such as video surveillance~\cite{surveillance15spletter} and dataset cleaning~\cite{dataclean04cce,dataclean04kdnet,dataclean05plos}.
For the application of dataset cleaning, outlier detection is usually used as a pre-processing step for the main tasks such as learning from open-set noisy labels~\cite{iterativeosnl18cvpr}, webly supervised learning~\cite{webly15iccv}, and open-set semi-supervised learning~\cite{openworldssl21arxiv}.
To construct an outlier detection benchmark on MNIST, one class should be chosen so that all samples that belong to this class are considered as inliers. A small fraction of samples from other classes are introduced as outliers to be detected. 

\smallskip
\keypoint{Evaluation}
Apart from F-scores, AUROC, and AUPR, the evaluation of outlier detectors can be also evaluated by the performance of the main task it supports. For example, if an outlier detector is used to purify a dataset with noisy labels, the performance of a classifier that is trained on the cleaned dataset can indicate the quality of the outlier detector.

\smallskip
\keypoint{Remark: On Inclusion of Outlier Detection} 
Interestingly, the outlier detection task can be considered as an outlier in the generalized OOD detection framework, since outlier detectors are operated on the scenario when all observations are given, rather than following the training-test scheme. Also, publications exactly on this topic are rarely seen in the recent deep learning venues. However, we still include outlier detection in our framework, because intuitively speaking, outliers also belong to one type of out-of-distribution, and introducing it can help familiarize readers more with various terms~(\eg OD, AD, ND, OOD) that have confused the community for a long while.

\subsection{Related Topics}
\label{sec:related_topics}
Apart from the five sub-topics that are described in our \emph{generalized OOD detection} framework (shown in Figure~\ref{fig:taxonomy}), we further briefly discuss five related topics below, which help clarify the scope of this survey.

\smallskip
\keypoint{Learning with Rejection (LWR)}
LWR~\cite{bartlett2008classification} can date back to early works on abstention~\cite{chow1970optimum, fumera2002support}, which considered simple model families such as SVMs~\cite{cortes1995support}.
The phenomenon of neural networks' overconfidence in OOD data is first revealed by~\cite{fooldnn15cvpr}. Despite methodologies differences, subsequent works developed on OOD detection and OSR share the underlying spirit of classification with the rejection option. 

\smallskip
\keypoint{Domain Adaptation/Generalization}
Domain Adaptation (DA)~\cite{dasurvey18neurocomp} and Domain Generalization (DG)~\cite{dgsurvey21arxiv} also follow ``open-world'' assumption. Different from generalized OOD detection settings, DA/DG expects the existence of covariate shift during testing without any semantic shift and requires classifiers to make accurate predictions into the same set of classes~\cite{liu2020open}.
Noticing that OOD detection commonly concerns detecting the semantic shift, which is complementary to DA/DG. In the case when both covariate and semantic shift take place, the model should be able to detect semantic shift while being robust to covariate shift. More discussion on relations between DA/DG and OOD detection is in Section~\ref{sec:future_direction}. 
The difference between DA and DG is that while the former requires extra but few training samples from the target domain, the latter does not.

\smallskip
\keypoint{Novelty Discovery} 
Novelty discovery~\cite{dtc19iccv,zhao2021novel,jia2021joint,vaze2022generalized,joseph2022novel} requires all observations to be given in advance as outlier detection does. The observations are provided in a semi-supervised manner, and the goal is to explore and discover the new categories and classes in the unlabeled set. Different from outlier detection where outliers are sparse, the unlabeled set in novelty discovery setting can mostly consist of, and even be overwhelmed by unknown classes.

\smallskip
\keypoint{Zero-shot Learning}
Zero-shot learning~\cite{zslsurvey19tist} has a similar goal of novelty discovery but follows the training-testing scheme. The test set is under the ``open-world'' assumption with unknown classes, which expects classifiers trained only on the known classes to perform classification on unknown testing samples with the help of extra information such as label relationships.

\smallskip
\keypoint{Open-world Recognition}
Open-world recognition~\cite{toopenworld15cvpr} aims to build a lifelong learning machine that can actively detect novel images~\cite{liu2019large}, label them as new classes, and perform continuous learning. It can be viewed as a combination of novelty detection (or open-set recognition) and incremental learning. 
\revise{More specifically, open-world recognition extends the concept of OSR by adding the ability to incrementally learn new classes over time. In open-world scenarios, the system not only identifies unknown instances but also can update its model to include these new classes as part of the known set. This approach is more dynamic and suited for real-world applications where the environment is not static, and new categories can emerge after the initial training phase~\cite{parmar2023open}.}

\smallskip
\keypoint{Conformal Prediction}
\revise{Conformal prediction (CP) stands as a robust statistical framework in machine learning, primarily designed to provide confidence measures for predictions~\cite{shafer2008tutorial,angelopoulos2021gentle}. Distinctively, it yields prediction intervals with specified confidence levels, transcending the limitations of mere point estimates.
In scenarios of OOD detection, the conformal prediction framework becomes particularly insightful: wider prediction intervals or lower confidence levels generated by conformal prediction methods can serve as indicators of such OOD data.
Although research at the intersection of CP and OOD detection is still emerging~\cite{kaur2022idecode,kaur2022codit,cai2021inductive}, the potential of applying the conformal prediction framework in this domain is significant and warrants further exploration.}

\subsection{Organization of Remaining Sections}
In this paper, we focus on the methodologies of OOD detection in Section~\ref{sec:ood}, providing a comprehensive overview of the different approaches that have been proposed in the literature. We also briefly introduce methodologies for other sub-tasks including AD, ND, OSR, and OD in Section~\ref{sec:others}, to provide readers with a broader understanding of OOD-related problems and inspire the development of more effective methods.
For each sub-task, we categorize and introduce the methodologies into four groups: 
\textbf{1) classification-based methods:} methods that largely rely on classifiers;
\textbf{2) density-based methods:} detecting OOD by modeling data density;
\textbf{3) distance-based methods:} using distance metrics (usually in the feature space) to identify OODs;
and \textbf{4) reconstruction-based methods:} methods featured by reconstruction techniques.
To offer readers further insights from an empirical perspective, we conduct a thorough analysis that provides a fair comparison between representative OOD detection methods and methods from other sub-tasks.
Additionally, we highlight some of the remaining problems and limitations that exist in the current generalized OOD detection field. We conclude this survey with a discussion on the open challenges and opportunities for future research. It is worth noting that a concurrent survey~\cite{salehi2021unified} provides a detailed explanation of OOD-related methods, which greatly complements our work.

\begin{figure*}[!t]
    \centering
    \includegraphics[width=\linewidth]{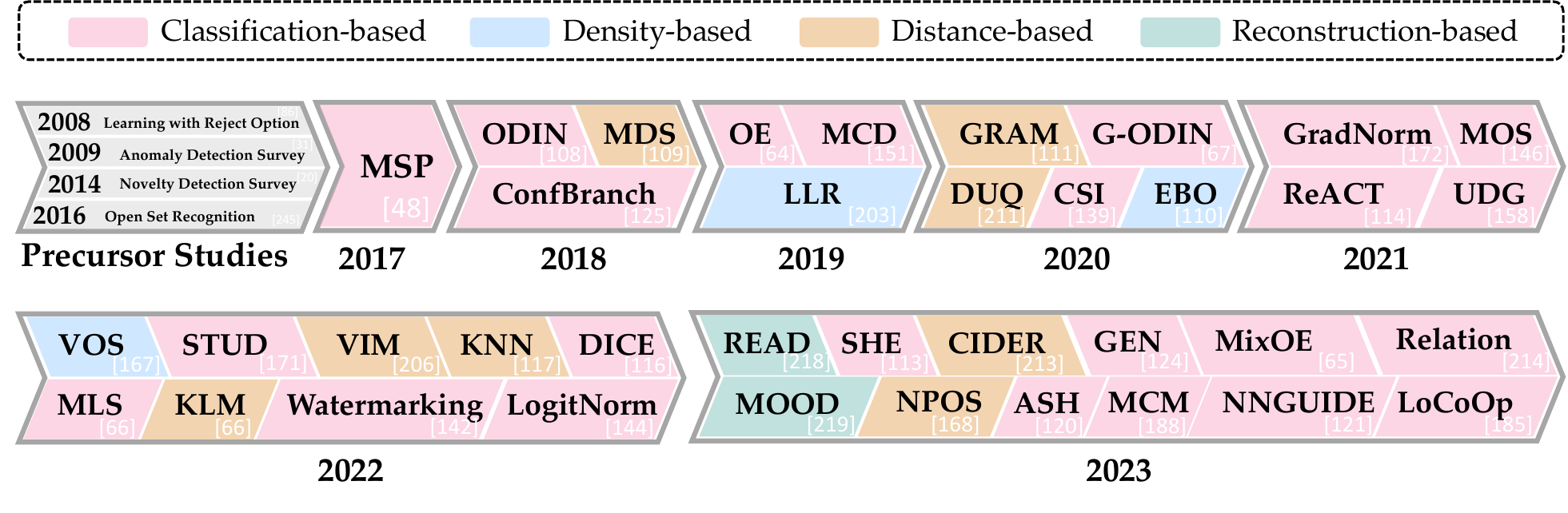}
    \vspace{-0.5cm}
    \caption{Timeline for representative OOD detection methodologies.
    Different colors indicate different categories of methodologies.
    Each method has its corresponding reference (inconspicuous white) in the lower right corner. Methods with high citations and open-source code are prioritized for inclusion in this figure.}

    \label{fig:timeline}
\end{figure*}

\begin{table*}[]
\caption{Paper list for out-of-distribution detection.}
\label{tab:method}
\centering
\begin{tabular}{@{}c|c|l|p{8cm}<{\centering}@{}}
\toprule
\multicolumn{3}{c|}{Sections} & References \\ \midrule
\midrule
\multicolumn{1}{c|}{\multirow{15}{*}{\begin{tabular}[c]{@{}c@{}} \textcolor{red}{$\S~$}\ref{sec:ood_classification} \\ Classification \end{tabular}}} &
\multicolumn{1}{c|}{\multirow{5}{*}{{\begin{tabular}[c]{@{}c@{}} \textcolor{red}{$\S~$}\ref{sec:ood_confcal} \\ Output-based \\ Methods \end{tabular}}}} &
\multirow{2}{*}{\begin{tabular}[c]{@{}c@{}} \textcolor{red}{a}:~Training-free
\end{tabular}}
& \cite{msp17iclr,odin18iclr,mahananobis18nips,energyood20nips,gram20icml,wang2021canmulti,she23iclr,sun2021tone,dong2022neural,sun2022dice,sun2022knn,mood21cvpr,gram19nipsw,she23iclr,djurisic2022extremely,park2023nearest,park2023understanding,jiang2023detecting,liu2023gen} \\  \cmidrule(l){3-4} 
&
\multicolumn{1}{l|}{}                              &
\multirow{3}{*}{\begin{tabular}[c]{@{}c@{}} \textcolor{red}{b}:~Training-based \end{tabular}}
& \cite{confbranch2018arxiv,wang2021energy,eloc18eccv,good20nips,aloe20arxiv,whyrelu19cvpr,blur20iclr, outliermining21ecml,ceda19cvpr,mixup19nips,cutmix19cvpr,cutout17arxiv,augmix19arxiv,hendrycks2021pixmix,csi20nips,ccu20arxiv,bibas2021single,wang2022watermarking,mood21cvpr,dongneural,godin20cvpr,wei2022mitigating,hierarchical18cvpr,mos21cvpr,hierarchyood,ksemantic18nips,nearood21arxiv,gan2021language} \\ \cmidrule(l){2-4} 
&
\multicolumn{1}{l|}{\multirow{3}{*}{{\begin{tabular}[c]{@{}c@{}} \textcolor{red}{$\S~$}\ref{sec:ood_confcal} \\ Outlier Exposure \\ \end{tabular}}}} &
\multirow{2}{*}{\begin{tabular}[c]{@{}c@{}} \textcolor{red}{a}:~Real Outliers
\end{tabular}}
& \cite{oe18nips,agnostophobia18nips,mcd19iccv,sina2020aaai, outliermining21ecml,abstention21arxiv,oecc21neurocomputing,outliermining21ecml, ming2022posterior,backgroundsampling20cvpr,Zhang_2023_WACV,pseudolabel20aaai,yang2021scood,lu2023uncertainty,lessbias19bmvc,katzsamuels2022training,wang2023learning} \\  \cmidrule(l){3-4} 
&
\multicolumn{1}{l|}{}                              &
\multirow{1}{*}{\begin{tabular}[c]{@{}c@{}} \textcolor{red}{b}:~Data Generation \end{tabular}}
& \cite{confcal18iclr,oodsg19nipsw,confgan18nipsw,maml20nips,du2022vos,npos2023iclr,wang2023out,zheng2023out,du2022unknown} \\ \cmidrule(l){2-4} 
&
\multicolumn{2}{l|}{
\textcolor{red}{$\S~$}\ref{sec:ood_gradient}:~Gradient-based Methods} & \cite{odin18iclr,huang2021importance,igoe2022useful}          \\
\cmidrule(l){2-4} 
&
\multicolumn{2}{l|}{
\textcolor{red}{$\S~$\ref{sec:ood_bayesian}}:~Bayesian Models}         
& \cite{mcdropout16icml,deepensemble17nips,practicalbnn19nips,dpn18nips,dpn19nips,dpn20nips,kim2021locally} \\
\cmidrule(l){2-4} 
&
\multicolumn{2}{l|}{
\textcolor{red}{$\S~$}\ref{sec:ood_foundation}:~OOD for Foundation Models}  & \cite{hendrycks2019using,nearood21arxiv,pretransformer20arxiv,ming2023does,miyai2023can,miyai2023locoop,lu2023likelihood,esmaeilpour2022zero,ming2022delving,wang2023clipn} \\
\cmidrule(l){1-4} 
\multicolumn{3}{c|}{
\multirow{2}{*}{\begin{tabular}[c]{@{}c@{}}
\textcolor{red}{$\S~$}\ref{sec:ood_density}:~Density-based Methods
\end{tabular}}}  
& \cite{dagmm18iclr,autogres19cvpr,gpnd18nips,adgan18ecml,advrec18cvpr,mahananobis18nips,flowreview20pami,residualflow20cvpr,glow18nips,pixelcnn16icml,jiang2021revisiting,dgmknow19nips,waic18arxiv,whyflow20nips,likelihoodratio19nips,ratiogm20iclr,vae20nips,haoqi2022vim} \\
\cmidrule(l){1-4} 
\multicolumn{3}{c|}{\textcolor{red}{$\S~$}\ref{sec:ood_distance}:~Distance-based Methods}               & \cite{mahananobis18nips,rmd21arxiv,cosinesim20accv,svae20eccv,onedim21cvpr,duq20icml,fss20arxiv, sun2022knn, ming2022cider,kim2023neural} \\ 
\cmidrule(l){1-4} 
\multicolumn{3}{c|}{\textcolor{red}{$\S~$}\ref{sec:ood_reconstruction}:~Reconstruction-based Methods}               & \cite{denouden2018improving,zhou2022rethinking,yang2022mask,jiang2022read,li2023mood}\\ 
\cmidrule(l){1-4} 
\multicolumn{3}{c|}{\textcolor{red}{$\S~$}\ref{sec:theoretical}:~Theoretical Analysis}               & \cite{zhang2021understanding,morteza2022provable,towardosr13pami,psvm14eccv,rudd2017extreme,pac18icml,bound21icml,fang2022out}\\ 
\bottomrule
\end{tabular}
\end{table*}

\section{OOD Detection: Methodology}
\label{sec:ood}
In this section, we introduce the methodology for OOD detection.
Initially, we explore classification-based models in Section~\ref{sec:ood_classification}. These models primarily utilize the model's output, such as softmax scores, to identify OOD instances. We further examine outlier exposure-based methods that leverage external data sources and other types of methods.
The later section is followed by density-based methods in Section~\ref{sec:ood_density}.
Distance-based methods will be introduced in Sections~\ref{sec:ood_distance}.
A brief discussion will be included at the end.

\subsection{Classification-based Methods}
\label{sec:ood_classification}
Research on OOD detection originated from a simple baseline, that is, using the maximum softmax probability as the indicator score of ID-ness~\cite{msp17iclr}. Early OOD detection methods focus on deriving improved OOD scores based on the output of neural networks.  

\subsubsection{Output-based Methods}
\label{sec:ood_confcal}
\keypoint{a.~Post-hoc Detection}
Post-hoc methods have the advantage of being easy to use without modifying the training procedure and objective. The property can be important for the adoption of OOD detection methods in real-world production environments, where the overhead cost of retraining can be prohibitive. Early work ODIN~\cite{odin18iclr} is a post-hoc method that uses temperature scaling and input perturbation to amplify the ID/OOD separability. Key to the method, a sufficiently large temperature has a strong smoothing effect that transforms the softmax score back to the logit space---which effectively distinguishes ID vs. OOD. Note that this is different from confidence calibration, where a much milder $T$ is employed. While calibration focuses on representing the true correctness likelihood of {ID data only}, the ODIN score is designed to maximize the gap between ID and OOD data and may no longer be meaningful from a predictive confidence standpoint. 
Built on the insights, recent work~\cite{energyood20nips, mood21cvpr} proposed using an energy score for OOD detection, which enjoys theoretical interpretation from a likelihood perspective~\cite{morteza2022provable}. Test samples with lower energy are considered ID and vice versa. 
JointEnergy score~\cite{wang2021canmulti} is then proposed to perform OOD detection for multi-label classification networks. The most recent work SHE~\cite{she23iclr} uses stored patterns that represent classes to measure the discrepancy of unseen data for OOD detection, which is hyperparameter-free and computationally efficient compared to classic energy methods.
Techniques such as layer-wise Mahalanobis distance~\cite{mahananobis18nips} and Gram Matrix~\cite{gram20icml} are implemented for better-hidden feature quality to perform density estimation.

Recently, one fundamental cause of the overconfidence issue on OOD data has been revealed that using mismatched BatchNorm statistics---that are estimated on ID data yet blindly applied to the OOD data in testing---can trigger abnormally high unit activations and model output accordingly~\cite{sun2021tone}. Therefore, ReAct~\cite{sun2021tone} proposes truncating the high activations, which establishes strong post-hoc detection performance and further boosts the performance of existing scoring functions. 
Similarly, NMD~\cite{dong2022neural} uses the activation means from BatchNorm layers for ID/OOD discrepancy.
While {ReAct} considers activation space, \cite{sun2022dice} proposes a weight sparsification-based OOD detection framework termed DICE. DICE ranks weights based on a measure of contribution and selectively uses the most salient weights to derive the output for OOD detection. By pruning away noisy signals, DICE provably reduces the output variance for OOD data, resulting in a sharper output distribution and stronger separability from ID data. 
\revise{In a similar vein, ASH~\cite{djurisic2022extremely} also targets the activation space but adopts a different strategy. It removes a significant portion (e.g., 90\%) of an input's feature representations from a late layer based on a top-K criterion, followed by adjusting the remaining activations (e.g., 10\%) either by scaling or assigning constant values, yielding surprisingly effective results.}

\keypoint{b.~Training-based Methods}
With the training phase, confidence can be developed via designing a confidence-estimating branch~\cite{confbranch2018arxiv} or class~\cite{wang2021energy}, ensembling with leaving-out strategy~\cite{eloc18eccv}, adversarial training~\cite{good20nips,aloe20arxiv,whyrelu19cvpr,blur20iclr, outliermining21ecml}, stronger data augmentation~\cite{ceda19cvpr,mixup19nips,cutmix19cvpr,cutout17arxiv,augmix19arxiv,hendrycks2021pixmix}, pretext training~\cite{csi20nips}, better uncertainty modeling~\cite{ccu20arxiv,bibas2021single}, input-level manipulation~\cite{odin18iclr,wang2022watermarking}, and utilizing feature or statistics from the intermediate-layer features~\cite{mood21cvpr,dongneural}. Especially, to enhance the sensitivity to covariate shift, some methods focus on the hidden representations in the middle layers of neural networks. Generalized ODIN, or G-ODIN~\cite{godin20cvpr} extended ODIN~\cite{odin18iclr} by using a specialized training objective termed DeConf-C and choosing hyperparameters such as perturbation magnitude on ID data. Note that we do not categorize G-ODIN as post-hoc method as it requires model retraining. Recent work~\cite{wei2022mitigating} shows that the overconfidence issue can be mitigated through Logit Normalization (LogitNorm), a simple fix to the common cross-entropy loss by enforcing a constant vector norm on the logits in training. Trained with LogitNorm, neural networks produce highly distinguishable confidence scores between in- and out-of-distribution data.

Some works redesign the label space to achieve good OOD detection performance. While commonly used to encode categorical information for classification, the one-hot encoding ignores the inherent relationship among labels. For example, it is unreasonable to have a uniform distance between \texttt{dog} and \texttt{cat} vs. \texttt{dog} and \texttt{car}.
To this end, several works attempt to use information in the label space for OOD detection.
Some works arrange the large semantic space into a hierarchical taxonomy of known classes~\cite{hierarchical18cvpr,mos21cvpr,hierarchyood}.
Under the redesigned label architecture, top-down classification strategy~\cite{hierarchical18cvpr,hierarchyood} and group softmax training~\cite{mos21cvpr} are demonstrated effective. 
Another set of works uses word embeddings to automatically construct the label space.
In \cite{ksemantic18nips}, the sparse one-hot labels are replaced with several dense word embeddings from different NLP models, forming multiple regression heads for robust training. When testing, the label, which has the minimal distance to all the embedding vectors from different heads, will be considered as the prediction. If the minimal distance crosses above the threshold, the sample would be classified as ``novel''.
Recent works further take the image features from language-image pre-training models~\cite{radford2021learning} to better detect novel classes, where the image encoding space also contains rich information from the language space~\cite{nearood21arxiv,gan2021language}. 

\subsubsection{Methods with Outlier Exposure}
\label{sec:ood_classification_oe}

\keypoint{a.~Real Outliers}
Another branch of OOD detection methods makes use of a set of collected OOD samples, or ``outlier'', during training to help models learn ID/OOD discrepancy.
Starting from the concurrent baselines that encourage a flat/high-entropic prediction on given OOD samples~\cite{oe18nips,agnostophobia18nips} and suppressing OOD feature magnitudes~\cite{agnostophobia18nips}, a follow-up work, MCD~\cite{mcd19iccv} uses a network with two branches, between which entropy discrepancy is enlarged for OOD training data. Another straightforward approach with outlier exposure spares an extra abstention (or rejection class) and considers all the given OOD samples in this class~\cite{sina2020aaai, outliermining21ecml, abstention21arxiv}.
A later work OECC~\cite{oecc21neurocomputing} noticed that an extra regularization for confidence calibration introduces additional improvement for OE.
To effectively utilize the given, usually massive, OOD samples, some work use outlier mining~\cite{outliermining21ecml, ming2022posterior} and adversarial resampling~\cite{backgroundsampling20cvpr} approaches to obtain a compact yet representative set.
\revise{In cases where the meaningful ``near''-OOD images are not available, MixOE \cite{Zhang_2023_WACV} proposes to interpolate between ID and ``far''-OOD images to obtain informative outliers for better regularization.} 
Other works consider a more practical scenario where given OOD samples contain ID samples, therefore using pseudo-labeling~\cite{pseudolabel20aaai} or ID filtering methods~\cite{yang2021scood} with optimal transport scheme~\cite{lu2023uncertainty} to reduce the interference of ID data.
In general, OOD detection with outlier exposure can reach a much better performance. 
However, research shows that the performance can be largely affected by the correlations between given and real OOD samples~\cite{lessbias19bmvc}. To address the issue, recent work~\cite{katzsamuels2022training} proposes a novel framework that enables effectively exploiting unlabeled in-the-wild data for OOD detection. Unlabeled wild data is frequently available since it is produced essentially for free whenever deploying an existing classifier in a real-world system. This setting can be viewed as training OOD detectors in their \emph{natural habitats}, which provide a much better match to the true test time distribution than data collected offline.

\keypoint{b.~Outlier Data Generation}
The outlier exposure approaches impose a strong assumption on the availability of OOD training data, which can be infeasible in practice. 
When no OOD sample is available, some methods attempt to synthesize OOD samples to enable ID/OOD separability.
Existing works leverage GANs to generate OOD training samples and force the model predictions to be uniform~\cite{confcal18iclr}, generate boundary samples in the low-density region~\cite{oodsg19nipsw}, or similarly, high-confidence OOD samples~\cite{confgan18nipsw}, or using meta-learning the update sample generation~\cite{maml20nips}.
However, synthesizing images in the high-dimensional pixel space can be difficult to
optimize. 
Recent work VOS~\cite{du2022vos} proposed synthesizing virtual outliers from the low-likelihood region in the feature space, which is more tractable given lower dimensionality. 
While VOS~\cite{du2022vos} is a parametric approach that models the feature space as a class-conditional Gaussian distribution, NPOS~\cite{npos2023iclr} also generates outlier ID data but in a non-parametric approach. \revise{Noticing the generated OOD data could be incorrect or irrelevant, DOE~\cite{wang2023out} synthesizes hard OOD data that leads to worst judgments to train the OOD detector with a min-max learning scheme, and ATOL~\cite{zheng2023out} uses auxiliary task to relieve the mistaken OOD generation.}
In object detection, \cite{du2022unknown} proposes synthesizing unknown objects from videos in the wild using spatial-temporal unknown distillation.

\subsubsection{Gradient-based Methods}
\label{sec:ood_gradient}
Existing OOD detection approaches primarily rely on the output (Section~\ref{sec:ood_classification}) or feature space for deriving OOD scores, while overlooking information from the gradient space. ODIN~\cite{odin18iclr} first explored using gradient information for OOD detection. In particular, ODIN proposed using input pre-processing by adding small perturbations obtained from the input gradients. The goal of ODIN perturbations is to increase the softmax score of any given input by reinforcing the model's belief in the predicted label. Ultimately the perturbations have been found to create a greater gap between the softmax scores of ID and OOD inputs, thus making them more separable and improving the performance of OOD detection. While ODIN only uses gradients {implicitly} through input perturbation, recent work proposed GradNorm~\cite{huang2021importance} which explicitly derives a scoring function from the {gradient space}. GradNorm employs the vector norm of gradients, backpropagated from the KL divergence between the softmax output and a uniform probability distribution.
\revise{A recent research~\cite{igoe2022useful} demonstrates that while gradient-based methods are effective, their success does not necessarily depend on gradients, but rather on the magnitude of learned feature embeddings and predicted output distribution.}

\subsubsection{Bayesian Models}
\label{sec:ood_bayesian}
A Bayesian model is a statistical model that implements Bayes' rule to infer all uncertainty within the model~\cite{jaynes1986bayesian}.
The most representative method is the Bayesian neural network~\cite{bnn12book}, which draws samples from the posterior distribution of the model via MCMC~\cite{mcmc06gamerman}, Laplace methods~\cite{laplace92cit,inbwnbnn20icmlw} and variational inference~\cite{meanfield89nn}, forming the epistemic uncertainty of the model prediction.
However, their obvious shortcomings of inaccurate predictions~\cite{howgoodbnn20icml} and high computational costs~\cite{objectbnn08ba} prevent them from wide adoption in practice.
Recent works attempt several less principled approximations including MC-dropout~\cite{mcdropout16icml} and deep ensembles~\cite{deepensemble00iwmcs,deepensemble17nips, maddox2019simple} for faster and better estimates of uncertainty. These methods are less competitive for OOD uncertainty estimation. 
Further exploration takes natural-gradient variational inference and enables practical and affordable modern deep learning training while preserving the benefits of Bayesian principles~\cite{practicalbnn19nips}.
Dirichlet Prior Network (DPN) is also used for OOD detection with an uncertainty modeling of three different sources of uncertainty: model uncertainty, data uncertainty, and distributional uncertainty, and form a line of works~\cite{dpn18nips,dpn19nips,dpn20nips}.
Recently, the Bayesian hypothesis test has been used to formulate OOD detection, with upweighting method and Hessian approximation for scalability~\cite{kim2021locally}.

\subsubsection{OOD Detection for Foundation Models}
\label{sec:ood_foundation}
\revise{
Foundation models~\cite{bommasani2021opportunities}, notably large-scale vision-language models~\cite{radford2021learning}, have demonstrated exceptional performance in a variety of downstream tasks. Their success is largely attributed to extensive pre-training on large-scale datasets. Several works~\cite{hendrycks2019using,nearood21arxiv,pretransformer20arxiv} reveal that well-pretrained models can significantly enhance OOD detection, particularly in challenging scenarios.
However, adapting (tuning) these models for downstream tasks with specific semantic (label) space in the training data remains a challenge, as simple approaches such as linear probing, prompt tuning~\cite{zhou2022coop,zhou2022cocoop,jia2022visual}, and adaptor-style fine-tuning methods~\cite{gao2023clip} do not have good results on OOD detection.
To advance the problem, a thorough investigation~\cite{ming2023does} examines how fine-tuned vision-language models are performed. Additionally, recent research~\cite{miyai2023can} highlights the impact of large-scale pretraining data and provides a systematic study on pretraining strategies on OOD detection performance.
On a technical front, LoCoOp~\cite{miyai2023locoop} introduces OOD regularization to a subset of CLIP's local features identified as OOD, enhancing prompt learning for better ID and OOD differentiation, and LSA~\cite{lu2023likelihood} uses a bidirectional prompt customization mechanism to enhance the image-text alignment.

The strong zero-shot learning capabilities of models like CLIP~\cite{radford2021learning} also open avenues for zero-shot OOD detection. This new setting aims to categorize known class samples and detect samples that do not belong to any of the known classes, where known classes are represented solely through textual descriptions or class names, eliminating the need for explicit training on these classes.
Addressing this, ZOC~\cite{esmaeilpour2022zero} trains a decoder based on CLIP's visual encoder to create candidate labels for OOD detection. While ZOC is computationally intensive and data-demanding, MCM~\cite{ming2022delving} opts for softmax scaling to align visual features with textual concepts for OOD detection.
A recent advancement, CLIPN~\cite{wang2023clipn}, innovatively integrates a ``no'' logic in OOD detection. Utilizing new prompts and a text encoder, along with novel opposite loss functions, CLIPN effectively tackles the challenge of identifying hard-to-distinguish OOD samples. This development marks a significant stride in enhancing the precision of OOD detection in complex scenarios.
}

\subsection{Density-based Methods}
\label{sec:ood_density}
Density-based methods in OOD detection explicitly model the in-distribution with some probabilistic models, and flag test data in low-density regions as OOD.
Although OOD detection can be different from AD in that multiple classes exist in the in-distribution, density estimation methods used for AD in Section~\ref{sec:anomaly} can be directly adapted to OOD detection by unifying the ID data as a whole~\cite{dagmm18iclr,autogres19cvpr,gpnd18nips,adgan18ecml,advrec18cvpr}.
When the ID contains multiple classes, class-conditional Gaussian distribution can explicitly model the in-distribution so that the OOD samples can be identified based on their likelihoods~\cite{mahananobis18nips}.
Flow-based methods~\cite{flowreview20pami,residualflow20cvpr,glow18nips,pixelcnn16icml,jiang2021revisiting} can also be used for probabilistic modeling.
While directly estimating the likelihood seems like a natural approach, some  works~\cite{dgmknow19nips,waic18arxiv,whyflow20nips} find that probabilistic models sometimes assign a higher likelihood for the OOD sample.
Several works attempt to solve the problems using likelihood ratio~\cite{likelihoodratio19nips}.
~\cite{ratiogm20iclr} finds that the likelihood exhibits a strong bias towards the input complexity and proposes a likelihood ratio-based method to compensate for the influence of input complexity.
Recent methods turn to new scores such as likelihood regret~\cite{vae20nips} or an ensemble of multiple density models~\cite{waic18arxiv}. 
To directly model the density of semantic space, SEM score is used with a simple combination of density estimation in the low-level and high-level space~\cite{yang2022fsood}.
Overall, generative models can be prohibitively challenging to train and optimize, and the performance can often lag behind the classification-based approaches (Section~\ref{sec:ood_classification}).

\subsection{Distance-based Methods}
\label{sec:ood_distance}
The basic idea of distance-based methods is that the testing OOD samples should be relatively far away from the centroids or prototypes of in-distribution classes. 
~\cite{mahananobis18nips} uses the minimum Mahalanobis distance to all class centroids for detection. A subsequent work splits the images into foreground and background and then calculates the Mahalanobis distance ratio between the two spaces~\cite{rmd21arxiv}. In contrast to the parametric approach, recent work~\cite{sun2022knn} shows strong promise of non-parametric nearest-neighbor distance for OOD detection. Unlike Mahalanobis, the non-parametric approach does not impose any distributional assumption about the underlying
feature space, hence providing stronger simplicity, flexibility, and generality.

For distance functions, some works use cosine similarity between test sample features and class features to determine OOD samples~\cite{cosinesim20accv,svae20eccv}.
The one-dimensional subspace spanned by the first singular vector of the training features is shown to be more suitable for cosine similarity-based detection~\cite{onedim21cvpr}. 
Moreover, other works leverage distances with radial basis function kernel~\cite{duq20icml}, Euclidean distance~\cite{fss20arxiv}, and geodesic distance~\cite{gomes2022igeood} between the input’s embedding and the class centroids.
Apart from calculating the distance between samples and class centroids, the feature norm in the orthogonal complement space of the principal space is shown effective on OOD detection~\cite{haoqi2022vim}. Recent work CIDER~\cite{ming2022cider} explores the usability of the embeddings in the hyperspherical space, where inter-class dispersion and inner-class compactness can be encouraged.

\subsection{Reconstruction-based Methods}
\label{sec:ood_reconstruction}
The core idea of reconstruction-based methods is that the encoder-decoder framework trained on the ID data usually yields different outcomes for ID and OOD samples. The difference in model performance can be utilized as an indicator for detecting anomalies.
For example, reconstruction models that are only trained by ID data cannot well recover the OOD data~\cite{denouden2018improving}, and therefore the OOD can be identified. While reconstruction-based models with pixel-level comparison seem not a popular solution in OOD detection for its expensive training cost, reconstructing with hidden features is shown as a promising alternative~\cite{zhou2022rethinking}.
Rather than reconstructing the entire image, recent work MoodCat~\cite{yang2022mask} masks a random portion of the input image and identifies OOD samples using the quality of the classification-based reconstruction results. READ~\cite{jiang2022read} combines inconsistencies from a classifier and an autoencoder by transforming the reconstruction error of raw pixels to the latent space of the classifier. \revise{MOOD~\cite{li2023mood} shows that masked image modeling for pretraining is beneficial to OOD detection tasks compared to contrastive training and classic classifier training.}

\subsection{Theoretical Analysis}
\label{sec:theoretical}
\revise{
Early theoretical research on OOD detection~\cite{zhang2021understanding} delves into the limitations of Deep Generative Models (DGMs) in OOD contexts. This work uncovers a critical flaw where DGMs frequently assign greater probabilities to OOD data compared to training data, attributing this issue primarily to model misestimation rather than the typical set hypothesis. This hypothesis posits that relevant out-distributions might be located in high-likelihood areas of the data distribution. The study concludes that any generalized OOD task must restrict the set of distributions that are considered out-of-distribution, as without any restrictions, the task is impossible.
Later work~\cite{morteza2022provable} advances the field by developing a comprehensive analytical framework aimed at enhancing theoretical understanding and practical performance of OOD detection methods in neural networks. Their innovative approach culminates in a novel OOD detection method that surpasses existing techniques in both theoretical robustness and empirical performance.

Another series of studies has been focused on Open-Set Learning (OSL). The seminal work in this domain~\cite{towardosr13pami} conceptualizes open-space risk for recognizing samples from unknown classes. The following research applies extreme value theory to OSL~\cite{psvm14eccv,rudd2017extreme}.
While probably approximately correct (PAC) theory is applied for OSR~\cite{pac18icml}, their method required test samples during training. Therefore, an investigation of the generalization error bound is conducted and proves the existence of a low-error OSL algorithm under certain assumptions~\cite{bound21icml}.
Still, under the PAC theory, a later study establishes necessary and sufficient conditions for the learnability of OOD detection in various scenarios~\cite{fang2022out}, including cases with overlapping and non-overlapping ID and OOD data. Their work also offers theoretical support for existing OOD detection algorithms and suggests that OOD detection is possible under certain practical conditions.

Despite these theoretical advancements, the field eagerly anticipates further research addressing aspects such as generalization in OOD detection, the explainability of these models, the integration of deep learning theory specific to OOD detection, and the exploration of foundation model theories pertinent to this area.
}

\subsection{Discussion}
\label{sec:ood_discussion}
The field of OOD detection has enjoyed rapid development since its emergence, with a large space of solutions. 
In the multi-class setting, the problem can be canonical to OSR (Section~\ref{sec:osr})---accurately classify test samples from ID within the class space $\mathcal{Y}$, and reject test samples with semantics outside the support of $\mathcal{Y}$. The difference often lies in the evaluation protocol. OSR splits a dataset into two halves: one set as ID and another set as OOD. In contrast, OOD allows a more general and flexible evaluation by considering test samples from different datasets or domains. 
Moreover, OOD detection encompasses a broader spectrum of learning tasks (\eg multi-label classification~\cite{wang2021canmulti}, object detection~\cite{du2022vos, du2022unknown}) and solution space. Apart from the methodology development, theoretical understanding has also received attention in the community~\cite{morteza2022provable}, providing provable guarantees and empirical analysis to understand how OOD detection performance changes with respect to data distributions.
\section{Methodologies from Other Sub-tasks}
\label{sec:others}
In this section, we briefly introduce methodologies for sub-tasks under the generalized OOD detection framework, including AD, ND, OSR, and OD, in hope that the methods from other sub-tasks can inspire more ideas for OOD detection community.

\subsection{Open Set Recognition}
\label{sec:osr}
The concept of OSR was first introduced in~\cite{towardosr13pami}, which showed the validity of 1-class SVM and binary SVM for solving the OSR problem. In particular, \cite{towardosr13pami} proposes the 1-vs-Set SVM to manage the open-set risk by solving a two-plane optimization problem instead of the classic half-space of a binary linear classifier. This paper highlighted that the open-set space should also be bounded, in addition to bounding the ID risk.

\smallskip
\keypoint{Classification-based Methods}
\label{sec:osr_classification}
Early works focused on logits redistribution using the compact abating probability (CAP)~\cite{cap14pami} and extreme value theory (EVT)~\cite{evt90appmath,evt12book,psvm14eccv}.
In particular, classic probabilistic models lack the consideration of open-set space. CAP explicitly models the probability of class membership abating from ID points to OOD points, and EVT focuses on modeling the tail distribution with extreme high/low values.
In the context of deep learning, OpenMax~\cite{openmax16cvpr} first implements EVT for neural networks. OpenMax replaces the softmax layer with an OpenMax layer, which calibrates the logits with a per-class EVT probabilistic model such as Weibull distribution.

To bypass open-set risk construction, some works attained good results without EVT. For example, some work uses a membership loss to encourage high activations for known classes, and uses large-scale external datasets to learn globally negative filters that can reduce the activations of novel images~\cite{dtl19cvpr}.
Apart from explicitly forcing discrepancy between known/unknown classes, 
other methods extract stronger features through an auxiliary task of transformation classification~\cite{gdfr20cvpr}, or mutual information maximization between the input image and its latent features~\cite{m2iosr21arxiv}, \etc.

Image generation techniques have been utilized to synthesize unknown samples from known classes, which helps distinguish between known vs. unknown samples~\cite{gopenmax17bmvc,osrci18eccv,proser21cvpr,kong2021opengan}.
While these methods are promising on simple images such as handwritten characters, they do not scale to complex natural image datasets due to the difficulty in generating high-quality images in high-dimensional space.
Another solution is to successively choose random categories in the training set and treat them as unknown, which helps the classifier to shrink the boundaries and gain the ability to identify unknown classes~\cite{collectdecision20tkde,onevsrest20arxiv}.
Moreover, \cite{intraclass19eusipco} splits the training data into typical and atypical subsets, which also helps learn compact classification boundaries.

\smallskip
\keypoint{Distance-based Methods}
\label{sec:osr_distance}
Distance-based methods for OSR require the prototypes to be class-conditional, which allows maintaining the ID classification performance.
Category-based clustering and prototyping are performed based on the visual features extracted from the classifiers. OOD samples can be detected by computing the distance \emph{w.r.t.} clusters~\cite{metric18bmvc,podn2020scireport}.
Some methods also leveraged contrastive learning to learn more compact clusters for known classes~\cite{peeler20cvpr,rpl20eccv}, which enlarge the distance between ID and OOD.
CROSR~\cite{crosr19cvpr} enhances the features by concatenating visual embeddings from both the classifier and reconstruction model for distance computation in the extended feature space.
Besides using features from classifiers, GMVAE~\cite{gmvae20aaai} extracts features using a reconstruction VAE, and models the embeddings of the training set as a Gaussian mixture with multiple centroids for the following distance-based operations.
Classifiers using nearest neighbors are also adapted for OSR problem~\cite{nndr17ml}. By storing the training samples, the nearest neighbor distance ratio is used for identifying unknown samples in testing.

\smallskip
\keypoint{Reconstruction-based Methods}
\label{sec:osr_reconstruction}
With similar motivations as Section~\ref{sec:ood_reconstruction},
reconstruction-based methods expect different reconstruction behavior for ID vs. OOD samples. The difference can be captured in the latent feature space or the pixel space of reconstructed images. 

By sparsely encoding images from the known classes, open-set samples can be identified based on their dense representation. Techniques such as sparsity concentration index~\cite{srosr16pami} and kernel null space methods~\cite{knda13cvpr,iknda17cvpr} are used for sparse encoding.

By fixing the visual encoder obtained from standard multi-class training to maintain ID classification performance, C2AE trains a decoder conditioned on label vectors and estimates the reconstructed images using EVT to distinguish unknown classes~\cite{c2ae19cvpr}.
Subsequent works use conditional Gaussian distributions by forcing different latent features to approximate class-wise Gaussian models, which enables classifying known samples as well as rejecting unknown samples~\cite{cgdl20cvpr}.
Other methods generate counterfactual images, which help the model focus more on semantics~\cite{counterfactual21cvpr}.
Adversarial defense is also considered in~\cite{osad20eccv} to enhance model robustness.

\smallskip
\keypoint{Discussion}
\label{sec:osr_hybrid}
Although there is not an independent section for density-based methods, these methods can play an important role and are fused as a critical step in some classification-based methods such as OpenMax~\cite{openmax16cvpr}. The density estimation on visual embeddings can effectively detect unknown classes without influencing the classification performance. A hybrid model also uses a flow-based density estimator to detect unknown samples~\cite{openhybrid20eccv}.

As introduced in Section~\ref{sec:tax_ood}, the general goal of OSR and OOD detection is aligned, that is to detect semantic shift from the training data. Therefore, we encourage methods from these two field should learn more from each other.
For example, apart from novel methods, OSR research also shows that a good classifier~\cite{vaze2021open} in the close-set is critical to OSR performance, which should also applicable to OOD detection tasks.


\subsection{Anomaly Detection \& Novelty Detection}
\label{sec:anomaly}
This section reviews methodologies for sensory and semantic AD and one-class ND. Notice that multi-classes ND is covered in the previous. 
Given homogeneous in-distribution data, approaches include density-based, reconstruction-based, distance-based, and hybrid methods. We also discuss theoretical works.

\smallskip
\keypoint{Density-based Methods}
\label{sec:ad_density}
Density-based methods model normal data (ID) distributions, assuming anomalous test data has low likelihood while normal data has higher likelihood. Techniques include classic density estimation, density estimation with deep generative models, energy-based models, and frequency-based methods.

Parametric density estimation assumes pre-defined distributions~\cite{anomalyparametric98pami}. Methods involve multivariate Gaussian distribution~\cite{mahalanobis00cils,mahalanobis18jesp}, mixed Gaussian distribution~\cite{gmm84siam,anomalygmm00icml}, and Poisson distribution~\cite{poisson16isi}. Non-parametric density estimation handles more complex scenarios~\cite{nonparametric91jasa} with histograms~\cite{histogram73cstm,histogram12wireless,histogram09traffic,hbos12ki} and kernel density estimation (KDE)~\cite{kde62math,kdeanomaly98ime,anomalykde18tkde}.

Neural networks generate high-quality features to enhance classic density estimation. Techniques include autoencoder (AE)~\cite{ae91aiche} and variational autoencoder (VAE)~\cite{vae13arxiv}-based models, generative adversarial networks (GANs)~\cite{gan14nips}, flow-based models~\cite{flows15icml,flowreview20pami}, and representation enhancement strategies.

EBMs use scalar energy scores to express probability density~\cite{energy11icml} and provide a solution for AD~\cite{energyad16icml}. Training EBMs can be computationally expensive, but score matching~\cite{scorematch05jmlr} and stochastic gradient Langevin dynamics~\cite{sgld11icml} enable efficient training.

Frequency domain analysis for AD includes methods like CNN kernel smoothing~\cite{highfreq20cvpr}, spectrum-oriented data augmentation~\cite{amplitude21iccv}, and phase spectrum targeting~\cite{phase21cvpr}. These mainly focus on sensory AD.

\smallskip
\keypoint{Reconstruction-based Methods}
\label{sec:ad_reconstruction}
These AD methods leverage model performance differences on normal and abnormal data in feature space or by reconstruction error.

Sparse reconstruction assumes normal samples can be accurately reconstructed using a limited set of basis functions, while anomalies have larger reconstruction costs and a dense representation~\cite{admm15jsps,adsparsecrowd17mta,adsparsevideo13tcsvt}. Techniques include $L_1$ norm-based kernel PCA~\cite{kpca13pr} and low-rank embedded networks~\cite{lren21aaai}.

Reconstruction-error methods assume a model trained on normal data will produce better reconstructions for normal test samples than anomalies. Deep models include AEs~\cite{recerrae18wts}, VAEs~\cite{recerrvae15ie}, GANs~\cite{ganad18iclr}, and U-Net~\cite{framepred18cvpr}.

AE/VAE-based models combine reconstruction-error with AE/VAE models~\cite{recerrae18wts,recerrvae15ie} and use strategies like reconstructing by memorized normality~\cite{menae19iccv,memad20cvpr}, adapting model architectures~\cite{rsr20iclr}, and partial/conditional reconstruction~\cite{scadn21aaai,gpnd18nips,conad19icml}. In semi-supervised AD, CoRA~\cite{cora19aaai} trains two AEs on inliers and outliers, using reconstruction errors for anomaly detection.
Reconstruction-error methods using GANs leverage the discriminator to calculate reconstruction error for anomaly detection~\cite{ganad18iclr}. Variants like denoising GANs~\cite{advrec18cvpr}, class-conditional GANs~\cite{ocgan19cvpr}, and ensembling~\cite{ganesb21aaai} further improve performance.
Gradient-based methods observe different patterns on training gradient between normalities and anomalies in a reconstruction task, using gradient-based representation to characterize anomalies~\cite{gradcon20eccv}.

\smallskip
\keypoint{Distance-based Methods}
\label{sec:ad_distance}
These methods detect anomalies by calculating the distance between samples and prototypes~\cite{wettschereck1994study}, requiring training data in memory. Methods include K-nearest Neighbors~\cite{tian2014anomaly} and prototype-based methods~\cite{munz2007traffic,syarif2012unsupervised}.

\smallskip
\keypoint{Classification-based Methods}
\label{sec:ad_classification}
AD and one-class ND are often formulated as unsupervised learning problems, but there are some supervised and semi-supervised methods as well. One-class classification (OCC) directly learns a decision boundary that corresponds to a desired density level set of the normal data distribution~\cite{occ02tax}. DeepSVDD~\cite{deepsvdd18icml} introduced classic OCC to the deep learning community. PU learning~\cite{pusurvey08isip,pusurvey20ml,pusurvey19iisa,deepsad20iclr} is proposed for the semi-supervised AD setting where unlabeled data is available in addition to the normal data. Self-supervised learning methods use pretext tasks such as contrastive learning~\cite{csi20nips}, image transformation prediction~\cite{goad20iclr,transform18nips}, and future frame prediction~\cite{ssmtl21cvpr}, where anomalies are more likely to make mistakes on the designed task.

One-class classification learns a decision boundary that corresponds to a desired density level set of the normal data distribution, which DeepSVDD~\cite{deepsvdd18icml} introduced to the deep learning community. PU learning~\cite{pusurvey08isip,pusurvey20ml,pusurvey19iisa,deepsad20iclr} is a popular method for the semi-supervised AD setting. Self-supervised learning methods use pretext tasks such as contrastive learning~\cite{csi20nips}, image transformation prediction~\cite{goad20iclr,transform18nips}, and future frame prediction~\cite{ssmtl21cvpr}, where anomalies are more likely to make mistakes on the designed task.

\smallskip
\keypoint{Discussion: Sensory \emph{vs} Semantic AD}
Sensory and semantic AD approaches assume the normal data as homogeneous, despite the presence of multiple categories within it. While semantic AD methods are mainly applicable to sensory AD problems, the latter can benefit from techniques that focus on lower-level features (e.g., flow-based and hidden feature-based), local representations, and frequency-based methods. Although current OOD detection tasks mostly focus on semantic shift, the method for Sensory AD might be especially helpful for far OOD detection, like ImageNet \vs Texture dataset.

\smallskip
\keypoint{Discussion: Theoretical Analysis}
In addition to algorithmic development, theoretical analysis of AD and one-class ND has also been provided in some works. For instance, \cite{pac18icml} constructs a clean set of ID and a mixed set of ID/OOD with identical sample sizes, achieving a PAC-style finite sample guarantee for detecting a certain portion of anomalies with the minimum number of false alarms. All these works could be beneficial to the theoretical works of OOD detection.

\subsection{Outlier Detection}
\label{sec:outlier}
Outlier detection (OD) observes all samples to identify significant deviations from the majority distribution. Though mostly studied in data mining, deep learning-based OD methods are used for data cleaning in open-set noisy data~\cite{iterativeosnl18cvpr,webly15iccv} and open-set semi-supervised learning~\cite{openworldssl21arxiv}.

\smallskip
\keypoint{Density-based Methods}
\label{sec:od_density}
OD methods include Gaussian distribution~\cite{std05bmj,devmedian13jesp}, Mahalanobis distance~\cite{mahalanobis00cils}, Gaussian mixtures~\cite{gmm09siam}, and Local outlier factor (LOF)~\cite{lof00sigmod}. RANSAC~\cite{fischler1981random} estimates parameters for a mathematical model. Classic density methods and NN-based density methods can also be applied.

\smallskip
\keypoint{Distance-based Methods}
\label{sec:od_distance}
Outliers can be detected by neighbor counting~\cite{distanceod13nips,distanceod10vldb}, DBSCAN clustering~\cite{dbscan96kdd}, and graph-based methods~\cite{knngraph04icpr,neibourgraph04jiis,largegraph10icml,graphad15dmkd,graphad03sigkdd,gbc07ictai,gbc12iccse,ngc21iccv,webly20acmmm}.

\smallskip
\keypoint{Classification-based Methods}
\label{sec:od_classification}
AD methods like Isolation Forest~\cite{isoforest08icdm} and OC-SVM~\cite{occ02tax,deepsvdd18icml} can be applied to OD. Deep learning models can identify outliers~\cite{li2017learning}. Techniques for robustness and feature generalizability include ensembling~\cite{nguyen2019self}, co-training~\cite{han2018co}, and distillation~\cite{li2017learning,webly20eccv}.

\smallskip
\keypoint{Discussion}
OD techniques are valuable for open-set semi-supervised learning, learning with open-set noisy labels, and novelty discovery. All these solutions can be applied especially when OOD samples are exposed during the training stage~\cite{yang2021scood}.

\begin{figure}[!t]
    \centering
    \includegraphics[width=0.9\linewidth]{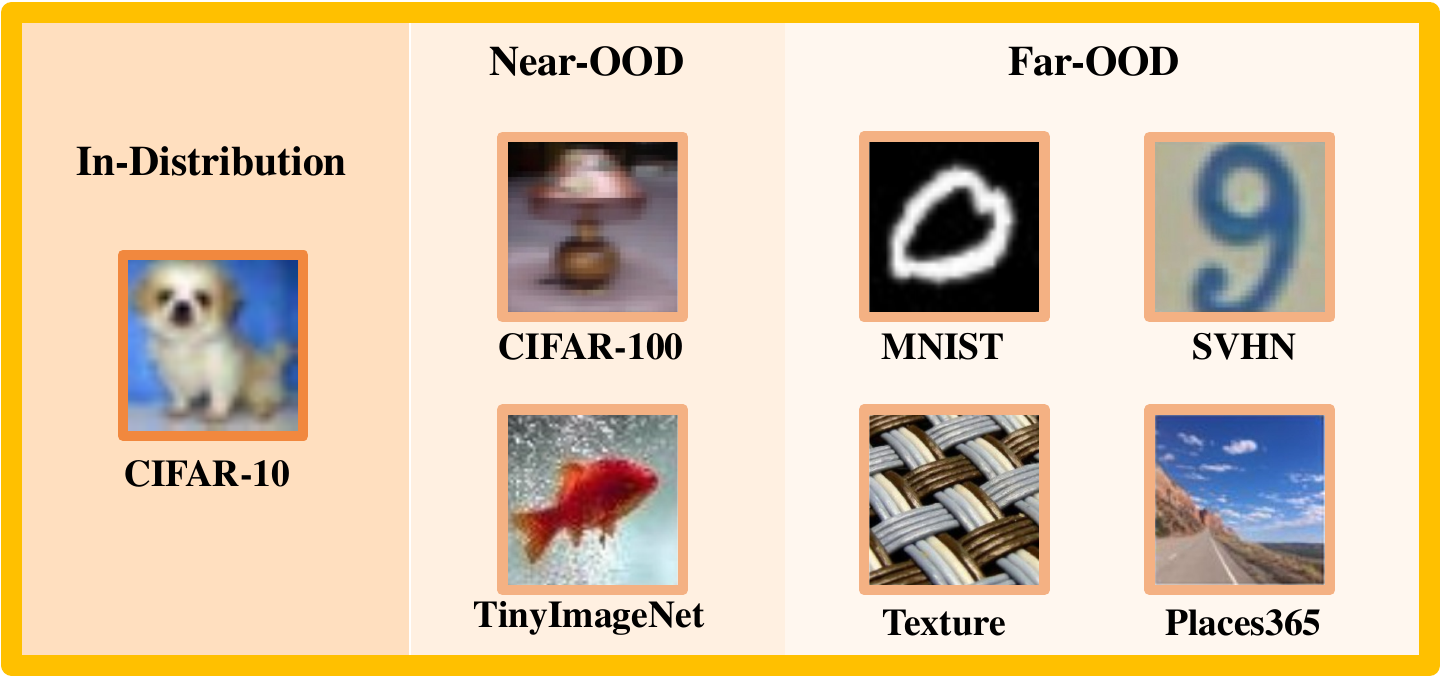}
    \caption{The illustration of CIFAR-10 benchmark that is used in Section~\ref{sec:benchmark}. The CIFAR-100 benchmark simply swaps the position of CIFAR-10 and CIFAR-100 in the figure.}
    \label{fig:benchmark}
\end{figure}
\begin{figure*}[!t]
    \centering
    \includegraphics[width=\linewidth]{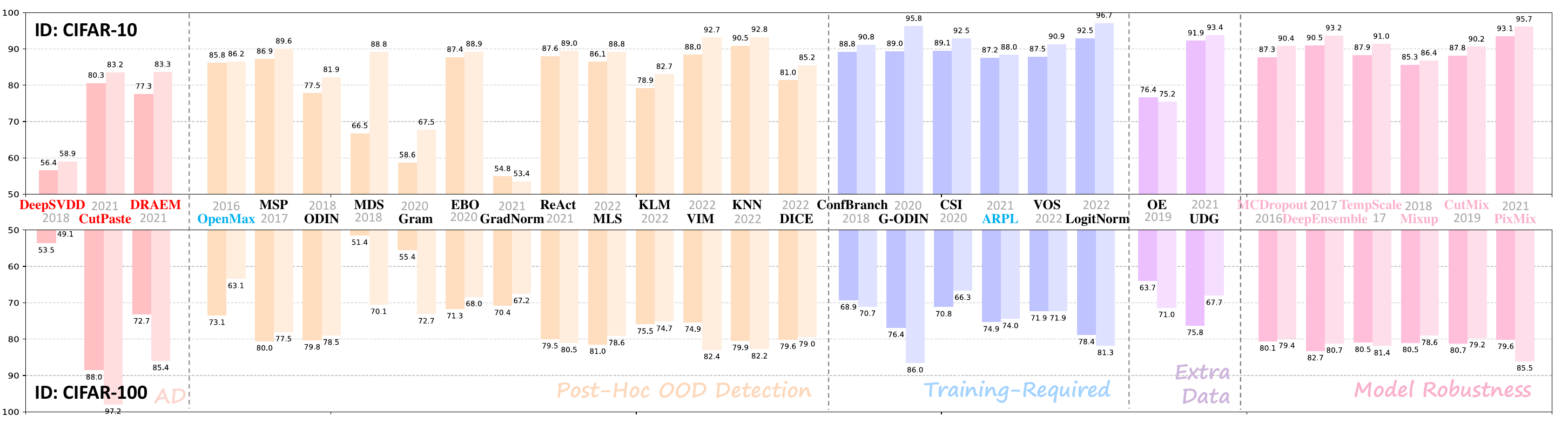}
    \vspace{-0.7cm}
\caption{Comparison between different methodologies under generalized OOD detection framework on the CIFAR-10/100 benchmarks. Results are from OpenOOD~\cite{yang2022openood}. Different colors denote the method categories. Each method reports near-OOD (left-bar) and far-OOD (right-bar) AUROC scores, as introduced in Section~\ref{sec:exp_metrics}. Method names in black originated for OOD detection, while in red are AD methods, blue for OSR methods, and pink for models from model uncertainty works.}
    \label{fig:exp}
\end{figure*}
\section{Benchmarks and Experiments}
\label{sec:benchmark}
In this section, we report the fair comparison of methodologies that from different categories on the CIFAR~\cite{cifar-10} benchmark. The report originated from OpenOOD benchmarks~\cite{yang2022openood}. We selected several popular AD methods, OOD detection methods (post-hot, training-required, and extra-data-required), and model robustness methods.

\subsection{Benchmarks and Metrics}
\label{sec:exp_metrics}
The common practice for building OOD detection benchmarks is to consider an entire dataset as in-distribution (ID), and then collect several datasets that are disconnected from any ID categories as OOD datasets. In this part, we show the results from two popular OOD benchmarks with ID datasets of CIFAR-10~\cite{cifar-10}, CIFAR-100~\cite{cifar-100} from OpenOOD (\cf Figure~\ref{fig:benchmark}), with each benchmark designing near-OOD and far-OOD datasets to facilitate detailed analysis of the OOD detectors. Near-OOD datasets only have semantic shift compared with ID datasets, while far-OOD further contains obvious covariate (domain) shift.

\noindent\textbf{CIFAR-10}\qquad
CIFAR-10~\cite{cifar-10} is a 10-class dataset for general object classification, which contains 50k training images and 10k test images. As for the OOD dataset, we construct near-OOD with CIFAR-100~\cite{cifar-100} and TinyImageNet~\cite{imagenet12nips}.
Notice that 1,207 images are removed from TinyImageNet since they actually belong to CIFAR-10 classes~\cite{yang2021scood}.
Far-OOD is built by MNIST~\cite{LeCun2005TheMD}, SVHN~\cite{svhn}, Texture~\cite{texture}, and Places365~\cite{zhou2017places} with 1,305 images are removed due to semantic overlaps.

\noindent\textbf{CIFAR-100}\qquad
Another OOD detection benchmark uses CIFAR-100~\cite{cifar-100} as an in-distribution, which contains 50k training images and 10k test images with 100 classes. 
For OOD dataset, near-OOD includes CIFAR-10~\cite{cifar-10} and TinyImageNet~\cite{tinyimages08pami}. 
Similar to the CIFAR-10 benchmark, 2,502 images are removed from TinyImageNet due to the overlapping semantics with CIFAR-100 classes~\cite{yang2021scood}.
Far-OOD consists of MNIST~\cite{LeCun2005TheMD}, SVHN~\cite{svhn}, Texture~\cite{texture}, and Places365~\cite{zhou2017places} with 1,305 images removed.

\noindent\textbf{Metrics}\qquad
We only report the AUROC scores, which measure the area under the Receiver Operating Characteristic (ROC) curve.

\subsection{Experimental Setup}
To ensure a fair comparison across methods that originate from different fields and have different implementations, unified settings with common hyperparameters and architecture choices are implemented. ResNet-18~\cite{he2016deep} is used as the backbone network. If the implemented method requires training, the widely accepted setting with SGD optimizer, a learning rate of 0.1, momentum of 0.9, and weight decay of 0.0005 for 100 epochs, is used. For further details, please refer to OpenOOD~\cite{yang2022openood,zhang2023openood}.

\subsection{Experimental Results and Findings}
\keypoint{Data Augmentation Methods are the Most Effective}
We split Figure~\ref{fig:exp} into several sections based on the method type. Generally, the most effective methods are those that use model uncertainty works with data augmentation techniques. This group mainly includes simple and effective methods such as preprocessing methods like PixMix~\cite{hendrycks2021pixmix} and CutMix~\cite{cutmix19cvpr}. PixMix achieves 93.1\% on Near-OOD in CIFAR-10, the best performance among all the methods in this benchmark. These methods also perform well in most of the other benchmarks. Similarly, other simple and effective methods to enhance model uncertainty estimation such as Ensemble~\cite{deepensemble00iwmcs} and Mixup~\cite{mixup19nips} also demonstrate excellent performance.

\keypoint{Extra Data Seems Not Necessary?}
Comparing UDG~\cite{yang2021scood} (the best from the extra-data part) with KNN~\cite{sun2022knn} (the best from the extra data-free part), we found that UDG's advantage is only in CIFAR-10 near-OOD, which is not satisfactory since a large quantity of real outlier data is required. In this benchmark, we use the entire TinyImageNet training set as the extra data, the choice of training outliers could greatly affect the performance of OOD detectors, so further exploration is needed.

\keypoint{Post-Hoc Methods Outperform Training in General}
Surprisingly, methods that require training do not necessarily perform better. In general, inference-only methods outperform trained methods. Nevertheless, the trained models can be generally used in conjunction with post-hoc methods, which could potentially further increase their performance.

\keypoint{Post-Hoc Methods are Making Progress}
In general, recent post-hoc methods have had better performance than previous methods since 2021, indicating that the direction of inference-only methods is promising and making progress. Recent methods show improvements in performance on more realistic datasets than previous methods, which focused on toy datasets. For example, the classic MDS performs well on MNIST but poorly on CIFAR-10 and CIFAR-100, while the recent KNN maintains good performance on MNIST, CIFAR-10, CIFAR-100, and also shows outstanding performance on ImageNet~\cite{yang2022openood}.

\keypoint{Some AD Methods are Good at Far-OOD}
Although anomaly detection (AD) methods were originally designed to detect pixel-level appearance differences on the MVTec-AD dataset, they have shown potency in far-OOD detection, such as with DRAEM and CutPaste. Both methods achieved high performance on far-OOD detection, especially when using CIFAR-100 as the in-distribution dataset.

\keypoint{Explore OpenOOD for More Experimental Findings}
\revise{Accompanying our survey, we lead the development of OpenOOD~\cite{yang2022openood}, an open-source codebase that provides a unified framework and benchmarking platform for conducting fair comparisons of various model architectures and OOD detection methods. OpenOOD is continuously updated and includes two comprehensive experimental reports~\cite{yang2022openood,zhang2023openood} that delve into extensive analysis and discovery\footnote{We also provide a \href{https://zjysteven.github.io/OpenOOD/}{leaderboard} to track SOTA methods.}. We encourage readers to explore OpenOOD's resources for a deeper understanding of key aspects such as selecting model architectures, utilizing pre-trained models, practical applications, and detailed implementation insights.}

\subsection{Exclusion of Covariate-Shift Detection}
\revise{While OpenOOD does not include settings for pure covariate shift, this was a deliberate choice. The primary focus is on semantic shifts, which are fundamental to OOD detection. By not separately analyzing covariate shifts, we aim to avoid potential misinterpretations and prevent the overemphasis on covariate shift detection. Experiments in~\cite{yang2022fsood} highlight a key finding: most current OOD detectors are more sensitive to covariate shifts than semantic shifts and lead to the concept of ``full-spectrum OOD detection'', advocating for models that \textbf{effectively generalize to handle covariate shifts} while \textbf{simultaneously detecting samples with semantic shifts}. More experimental evaluations can be found in OpenOOD v1.5~\cite{zhang2023openood}.}

\section{Challenges and Future Directions}
\label{sec:future}
In this section, we discuss the challenges and future directions of generalized OOD detection. 

\subsection{Challenges}
\keypoint{a.~Proper Evaluation and Benchmarking}
We hope this survey can clarify the distinctions and connections of various sub-tasks, and help future works properly identify the target problem and benchmarks within the framework. The mainstream OOD detection works primarily focus on detecting semantic shifts.
Admittedly, the field of OOD detection can be very broad due to the diverse nature of distribution shifts. 
Such a broad OOD definition also leads to some challenges and concerns~\cite{semanticanomaly20aaai,gan2021language}, which advocate a clear specification of OOD type in consideration (\eg semantic OOD, adversarial OOD, \etc) so that proposed solutions can be more specialized.
Besides, the motivation of detecting a certain distribution shift also requires clarification. While rejecting classifying samples with semantic shift is apparent, detecting sensory OOD should be specified to some meaningful scenarios to contextualize the necessity and practical relevance of the task. 

We also urge the community to carefully construct the benchmarks and evaluations. It is noticed that early work~\cite{msp17iclr} ignored the fact that some OOD datasets may contain images with ID categories, causing inaccurate performance evaluation. 
Fortunately, recent OOD detection works~\cite{yang2021scood}
have realized this flaw and pay special attention to removing ID classes from OOD samples to ensure proper evaluation. 

\keypoint{b.~Outlier-free OOD Detection}
The outlier exposure approach~\cite{oe18nips} imposes a strong assumption of the availability of  OOD  training data,  which can be difficult to obtain in practice. Moreover, one needs to perform careful de-duplication to ensure that the outlier training data does not contain ID data. These restrictions may lead to inflexible solutions and prevent the adoption of methods in the real world. Going forward, a major challenge for the field is to devise outlier-free learning objectives that are less dependent on auxiliary outlier dataset. 

\keypoint{c.~Tradeoff Between Classification and OOD Detection}
In OSR and OOD detection, it is important to achieve the dual objectives simultaneously: one for the ID task (\eg image classification), another for the OOD detection task. For a shared network, an inherent trade-off may exist between the two tasks. Promising solutions should strive for both. These two tasks may or may not contradict each other, depending on the methodologies. For example, ~\cite{liu2019large} advocated the integration of image classification and open-set recognition so that the model will possess the capability of discriminative recognition on known classes and sensitivity to novel classes at the same time.
~\cite{vaze2021open} also showed that the ability of detecting novel classes can be highly correlated with its accuracy on the closed-set classes.
~\cite{yang2021scood} demonstrated that optimizing for the cluster compactness of ID classes may facilitate both improved classification and distance-based OOD detection performance. Such solutions may be more desirable than ND, which develops a binary OOD detector separately from the classification model, and requires deploying two models. 

\keypoint{d.~Real-world Benchmarks and Evaluations}
Current methods in OOD detection are predominantly evaluated on smaller datasets like CIFAR. However, it has been observed that strategies effective on CIFAR may not perform as well on larger datasets like ImageNet, which has a more extensive semantic space. This discrepancy underscores the importance of conducting OOD detection evaluations in large-scale, real-world settings. Consequently, we recommend future research to focus on benchmarks based on ImageNet for OOD detection~\cite{mos21cvpr} and to explore large-scale Open Set Recognition (OSR) benchmarks~\cite{vaze2021open} to fully test the effectiveness of these methods. \revise{Additionally, recent research~\cite{bitterwolf2023or} highlights the presence of erroneous samples in ImageNet OOD benchmarks and introduces the corrected NINCO dataset for more accurate evaluations. Furthermore, expanding the scope of benchmarks to encompass real-world scenarios, such as more realistic datasets~\cite{koh2021wilds,cultrera2023leveraging}, and object-level OOD detection~\cite{du2022vos,du2022unknown}, can provide valuable insights, especially in safety-critical applications like autonomous driving.}

\subsection{Future Directions}
\label{sec:future_direction}

\keypoint{a.~Methodologies across Sub-tasks}
Due to the inherent connections among different sub-tasks, their solution space can be shared and inspired by each other. For example, the recent emerging density-based OOD detection research (\cf Section~\ref{sec:ood_density}) can draw insights from the density-based AD methods (\cf Section~\ref{sec:ad_density}) that have been around for a long time.

\keypoint{b.~OOD Detection \& Generalization}
An open-world classifier should consider two tasks,
\ie being robust to covariate shift while being aware of the semantic shift. Existing works pursue these two goals independently. Recent work proposes a semantically coherent OOD detection framework~\cite{yang2021scood} that encourages detecting semantic OOD samples while being robust to negligible covariate shift. 
Given the vague definition of OOD, \cite{ming2022spurious} proposed a formalization of OOD detection by explicitly taking into account the separation
between invariant features (semantically related) and environmental features (non-semantic). The work highlighted that spurious environmental features in the training set can significantly impact
OOD detection, especially when the semantic OOD data contains the spurious feature. 
Further, full-spectrum OOD detection~\cite{yang2022fsood} highlights the effects of ``covariate-shifted in-distribution'', and show that most of the previous OOD detectors are unfortunately sensitive to covariate shift rather than semantic shift. This setting explicitly promotes the generalization ability of OOD detectors.
Recent works on open long-tailed recognition~\cite{liu2019large}, open compound domain adaptation~\cite{liu2020open}, open-set domain adaptation~\cite{panareda2017open} and open-set domain generalization~\cite{shu2021open} consider the potential existence of open-class samples.
Looking ahead, we envision great research opportunities on how OOD detection and OOD generalization can better enable each other~\cite{liu2019large}, in terms of both algorithmic design and comprehensive performance evaluation.

\keypoint{c.~OOD Detection \& Open-Set Noisy Labels}
Existing methods of learning from open-set noisy labels focus on suppressing the negative effects of noise~\cite{iterativeosnl18cvpr,mopro20iclr}. However,
the open-set noisy samples can be useful for outlier exposure (\cf ~Section \ref{sec:ood_classification_oe})~\cite{ngc21iccv} and potentially benefit OOD detection.
With a similar idea, the setting of open-set semi-supervised learning can be promising for OOD detection.
We believe the combination of OOD detection and the previous two fields can provide more insights and possibilities.

\keypoint{d.~OOD Detection For Broader Learning Tasks}
As mentioned in Section~\ref{sec:ood_discussion}, OOD detection encompasses a broader spectrum of learning tasks, including multi-label classification~\cite{wang2021canmulti}, object detection~\cite{du2022vos, du2022unknown}, image segmentation~\cite{hendrycks2019scaling}, time-series prediction~\cite{kaur2022codit}, and LiDAR-based 3D object detection~\cite{nguyen2022out}.
For the classification task itself, the researchers also extended the OOD detection technique to improve the reliability of zero-shot pretrained models~\cite{esmaeilpour2022zero} (\eg CLIP).
Furthermore, some studies focus on applying OOD detection methods to produce reliable image captions~\cite{shalev2022baseline}. 
\revise{Recent advancements extend OOD detection to continuously adaptive or online learning environments~\cite{wu2023meta}. Additionally, OOD detection shows promise in addressing model reliability issues in broader applications, like mitigating hallucination problems in large language models~\cite{zhou2020detecting}. The integration of OOD detection methods promises to enhance the reliability and practicality of models across various fields, and insights from these fields could, in turn, further refine OOD detection techniques.}

\keypoint{e.~OOD Detection with World Models}
\revise{The existing works utilizing foundation models, particularly multi-modal ones such as CLIP~\cite{radford2021learning}, have significantly enhanced OOD detection performance, as discussed in Section~\ref{sec:ood_foundation}.
Starting from this, recent advancements have further focused on leveraging the extensive world knowledge encapsulated in Large Language Models~\cite{dai2023exploring}. This approach aligns with the rapid development in multi-modal world models~\cite{yang2023dawn,liu2023visual,li2023otter}, presenting burgeoning opportunities for further innovation within the OOD detection community.
}

\section{Conclusion}
\label{sec:conclusion}
In this survey, we comprehensively review five topics: AD, ND, OSR, OOD detection, and OD, and unify them as a framework of \emph{generalized OOD detection}. By articulating the motivations and definitions of each sub-task, we encourage follow-up works to accurately locate their target problems and find the most suitable benchmarks.
By sorting out the methodologies for each sub-task, we hope that readers can easily grasp the mainstream methods, identify suitable baselines, and contribute future solutions in light of existing ones.
By providing insights, challenges, and future directions, we hope that future works will pay more attention to the existing problems and explore more interactions across other tasks within or even outside the scope of generalized OOD detection.
\section*{Acknowledgment}
This study is supported by the Ministry of Education, Singapore, under its MOE AcRF Tier 2 (MOE-T2EP20221- 0012), NTU NAP, and under the RIE2020 Industry Alignment Fund – Industry Collaboration Projects (IAF-ICP) Funding Initiative, as well as cash and in-kind contribution from the industry partner(s).
YL is supported by the Office of the Vice Chancellor for Research and Graduate Education (OVCRGE) with funding from the Wisconsin Alumni Research Foundation (WARF).

\section*{Data Availability Statement}
The datasets analyzed during the current study in Section~\ref{sec:benchmark} are available in the OpenOOD repository, https://github.com/Jingkang50/OpenOOD.


    
    

    



\bibliographystyle{ieeetr}
\bibliography{citation}   

\end{document}